%% file: iclr2026_conference.tex
\documentclass{article} 
\usepackage{iclr2026_conference,times}

\input{math_commands.tex}

\usepackage[colorlinks, linkcolor=red, anchorcolor=darkblue, citecolor=mygreen2, pagebackref]{hyperref}
\usepackage{url}

\usepackage{natbib}
\usepackage{color}
\usepackage{graphicx}
\definecolor{DeepPink}{RGB}{255,20,147}
\usepackage{booktabs, multirow}
\usepackage{xcolor,amsmath}
\definecolor{darkblue}{rgb}{0,0.08,0.45}
\definecolor{cvprblue}{rgb}{0.21,0.49,0.74}
\definecolor{mygreen2}{RGB}{0 205 0}
\usepackage{colortbl}
\usepackage{amsfonts,bm,pifont}
\usepackage{wrapfig}
\usepackage{marvosym}
\usepackage{placeins}

\title{
\raisebox{-0.15\height}{\includegraphics[height=0.6cm]{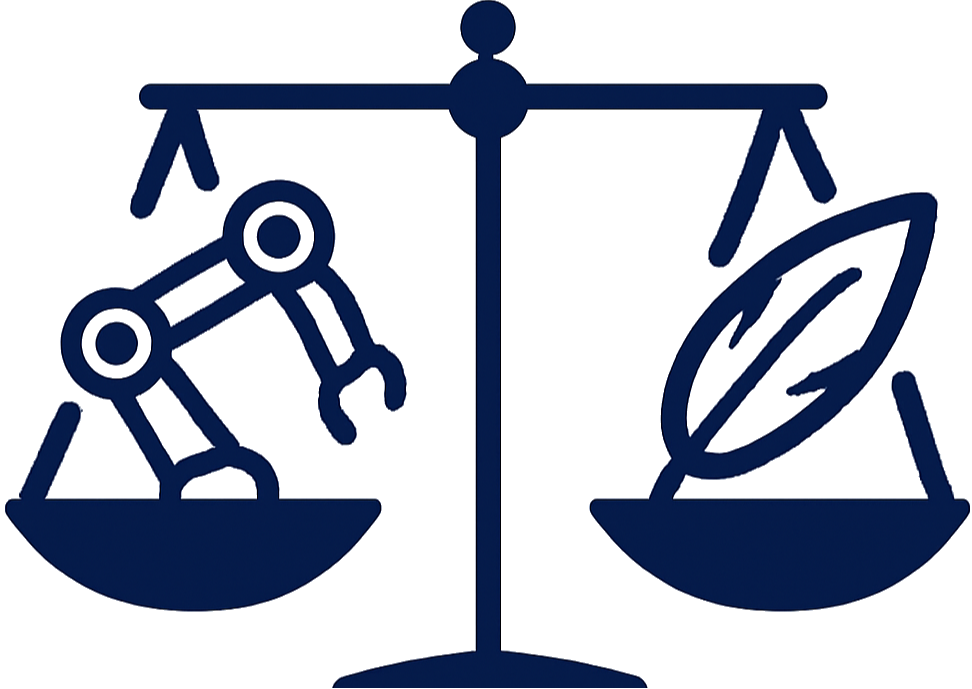}}\hspace{0.25em}
VLA-Adapter: An Effective Paradigm for Tiny-Scale Vision-Language-Action Model}

\author{Yihao Wang$^{1,2,4,*,\diamondsuit}$ \quad Pengxiang Ding$^{2,3,4,*,\dagger}$ \quad Lingxiao Li$^{1,4,5}$ \quad Can Cui$^{2,4}$\\ 
\textbf{Zirui Ge$^{3,4}$ \quad Xinyang Tong$^{2,4}$ \quad Wenxuan Song$^{4,6}$ \quad Han Zhao$^{2,3,4}$ \quad Wei Zhao$^{2,4}$}\\ 
\textbf{Pengxu Hou$^{6}$ \quad Siteng Huang$^{2}$ \quad Yifan Tang$^{1}$ \quad Wenhui Wang$^{1}$ \quad Ru Zhang$^{1,}$\textsuperscript{\Letter}}\\
\textbf{Jianyi Liu$^{1}$ \quad Donglin Wang$^{2,}$\textsuperscript{\Letter}}\\
$^{1}$Beijing University of Posts and Telecommunications \thinspace $^{2}$Westlake University \thinspace $^{3}$Zhejiang University\\
$^{4}$OpenHelix Team \quad $^{5}$State Key Laboratory of Networking and Switching Technology\\
$^{6}$The Hong Kong University of Science and Technology (Guangzhou)\\
$^{*}$Equal contribution: yh-wang@bupt.edu.cn; \thinspace dingpx2015@gmail.com\\
\textsuperscript{\Letter}Corresponding Author \thinspace 
$^{\dagger}$Project Lead \thinspace
$^{\diamondsuit}$Work done during interning at Westlake University\\
}

\iclrfinalcopy 
\begin{document}

\maketitle

\begin{abstract}
    Vision-Language-Action (VLA) models typically bridge the gap between perceptual and action spaces by pre-training a large-scale Vision-Language Model (VLM) on robotic data. While this approach greatly enhances performance, it also incurs significant training costs. In this paper, we investigate how to effectively bridge vision-language (VL) representations to action (A). We introduce VLA-Adapter, a novel paradigm designed to reduce the reliance of VLA models on large-scale VLMs and extensive pre-training. To this end, we first systematically analyze the effectiveness of various VL conditions and present key findings on which conditions are essential for bridging perception and action spaces. Based on these insights, we propose a lightweight Policy module with Bridge Attention, which autonomously injects the optimal condition into the action space. In this way, our method achieves high performance using only a 0.5B-parameter backbone, without any robotic data pre-training. Extensive experiments on both simulated and real-world robotic benchmarks demonstrate that VLA-Adapter not only achieves state-of-the-art level performance, but also offers the fast inference speed reported to date. Furthermore, thanks to the proposed advanced bridging paradigm, VLA-Adapter enables the training of a powerful VLA model in just 8 hours on a single consumer-grade GPU, greatly lowering the barrier to deploying the VLA model. Project page: \textcolor{DeepPink}{\url{https://vla-adapter.github.io/}}.
\end{abstract}

\input{Sections/Introduction}
\input{Sections/Related_work}
\input{Sections/Methodology}
\input{Sections/Experiments}
\input{Sections/Conclusion}
\input{Sections/Limitation}
\input{Sections/Acknowledgments}

\bibliography{iclr2026_conference}
\bibliographystyle{iclr2026_conference}

\newpage
\appendix
\input{Sections/Appendix}

\end{document}

%% file: math_commands.tex
\usepackage{amsmath,amsfonts,bm}

\def\eqref#1{equation~\ref{#1}}

\def\1{\bm{1}}

\DeclareMathAlphabet{\mathsfit}{\encodingdefault}{\sfdefault}{m}{sl}
\SetMathAlphabet{\mathsfit}{bold}{\encodingdefault}{\sfdefault}{bx}{n}



%% file: Sections/Introduction.tex
\section{Introduction}\label{Section_introduction}

\begin{wrapfigure}{r}{9cm}
	\centering
	\vspace{-0.3cm}
	\includegraphics[width=0.6\textwidth]{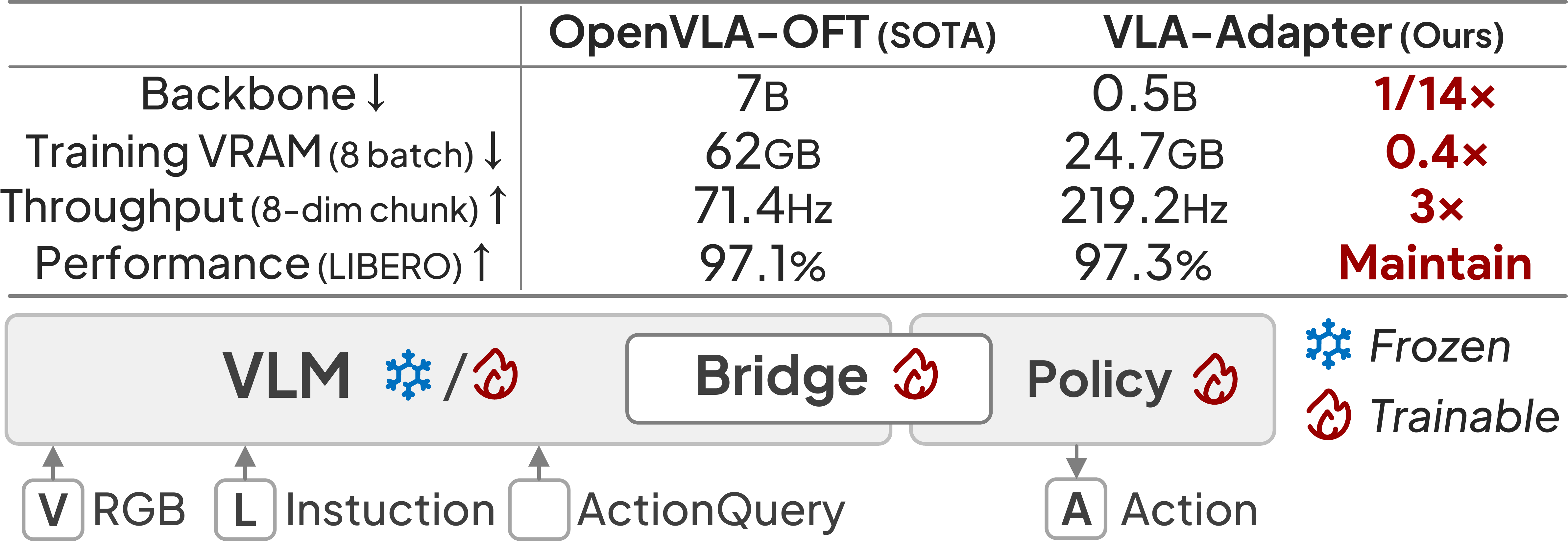}
	\vspace{-0.1cm}
	\caption{Characteristics of VLA-Adapter. ``$\downarrow$" is that smaller values are better, and vice versa. Our paradigm can effectively obtain the SOTA-level VLA model using a tiny-scale backbone.}\label{Figure_teaser}
	\vspace{-0.25cm}
\end{wrapfigure}

In the past two years, with significant breakthroughs in multimodal LLMs \citep{Prismatic-2024, PaliGemma-2024, LLaVA-2023, Eagle-2-2025}, developing robot systems with general perception, understanding, and behavior capabilities has become a key research direction in artificial intelligence. In particular, the emergence of the Vision-Language-Action (VLA) model offers a new solution for enabling robot operations driven by instructions \citep{OpenVLA-2024, OpenHelix-2025, OpenVLA-OFT-2025, PDVLA-2025, WorldVLA-2025, DreamVLA-2025, MemoryVLA-2025}. Research on VLA primarily focuses on extracting multimodal information and aligning it with the action space to generate the high-quality actions \citep{Octo-2024, RDT-2025, FlowVLA-2025, LongVLA-2025}.

Current VLA models typically require large-scale embodied data (e.g., Open X-Embodiment \citep{OpenX-2024} and DROID \citep{DROID-2024}) to pre-train Multimodal Large Language Models (MLLMs) (Especially, VLMs) for task adaptability \citep{GR-2-2024}, which is then passed to the designed Policy network \citep{RoboDual-2024, RoboFlamingo-2024} to decode or generate actions for handling the tasks in the diverse environments \citep{LIBERO-2023, CALVIN-2022}. 

However, when confronted with high-dimensional control environments, VLA models still face several bottlenecks, including reliance on large-scale VLMs, slow fine-tuning speed, high GPU memory (VRAM) consumption, and low inference efficiency (throughput), as shown in Figure \ref{Figure_teaser}. To this end, it is necessary to explore the most essential but rarely discussed question in the VLA field: \textbf{\textit{How to bridge the gap of VL (vision-language representations) to A (action) more effectively?}}

To answer this question, we propose VLA-Adapter, a novel bridging paradigm for VLA. We systematically explore how different conditions influence action generation and give some key findings for VLA design. On this basis, we built a Policy network with Bridge Attention to autonomously inject the optimal condition into the action space. Experiments show that VLA-Adapter has superior performance, high inference efficiency, and fast throughput with a tiny-scale backbone. It significantly lowers the barrier to VLA deployment. The main contributions are summarized as follows.

\begin{itemize}
	\item To our knowledge, this work is the first systematic analysis of bridging paradigms' effects on action generation. And we also give some key findings of the VLA model design.
	\item VLA-Adapter transfers the sufficient multimodal information to the proposed Policy Network for action generation, effectively bridging the modality gap from VL to A. 
	\item Rich experiments show that VLA-Adapter has a higher success rate, smaller scale, lower tuning cost, and faster inference in diverse simulated and real-world robotic tasks.
\end{itemize}

%% file: Sections/Related_work.tex
\section{Related Work}\label{Section_related_work}

\subsection{Vision-Language-Action (VLA) Models} \label{Subsection_VLA}
Recently, leveraging pre-trained Vision-Language Models (VLMs) \citep{Prismatic-2024, PaliGemma-2024, LLaVA-2023, Eagle-2-2025} to control robots for performing various daily tasks has substantially accelerated research in embodied intelligence. This has emerged as a prominent research focus \citep{pi0-2025, pi0.5-2025, SmolVLA-2025, GR00T-N1-2025, RDT-2025, Being-H0-2025, GR3-2025, G0-2025, QUAR-2024, CLOVER-2024, LongVLA-2025, QUARTOnline-2025}. These models are referred to as the VLA models.

Typically, VLA models require large-scale embodied datasets, such as Open X-Embodiment \citep{OpenX-2024}, for pre-training \citep{RDT-2025, GR3-2025}. This process integrates VLMs with a dedicated Policy network \citep{Hume-2025, RoboFlamingo-2024}, allowing the system to decode or generate action sequences for diverse tasks in an end-to-end manner. Moreover, dual-system VLA architectures \citep{LCB-2024, RoboDual-2024, OpenHelix-2025} have recently garnered attention. These methods typically introduce an intermediate latent token to connect the VLMs and the Policy, using an asynchronous mechanism to enhance coordination between the two systems \citep{HiRT-2024}. This design mitigates latency issues during action generation.

Consequently, how to effectively and efficiently bridge the gap from the vision-language perception space to the action space has become a key challenge in the design of VLA models.

\subsection{Bridging from Perception to Action Space} \label{sec22} 
Earlier studies \citep{OpenVLA-2024, RT-2-2023, RT-1-2023} attempted to directly align perception and action spaces by discretizing actions into tokens. However, this discretization inevitably introduces inherent loss. Recent studies have shifted their focus toward continuous action spaces \citep{RoboVLMs-2024, GR00T-N1-2025, pi0-2025, SmolVLA-2025, OpenVLA-OFT-2025}. Based on the types of perceptual features utilized to bridge to the action space, they can be categorized:

\paragraph{1) Raw Features from VLMs.} Raw features (refer to vision and language representations) are extracted directly from the VLM. Early methods extract representations from the final-layer VLM, operating under the assumption that it encodes the most task-relevant semantic information \citep{RoboVLMs-2024, HiRT-2024}. More recent methods leverage the intermediate-layer features within the VLM \citep{pi0-2025}. They believe that such representations may retain richer multimodal information, thereby benefiting Policy in tasks that demand fine-grained perception or complex reasoning. For example, some studies use features from a middle layer \citep{GR00T-N1-2025}, the first-half layers \citep{SmolVLA-2025}, or all intermediate-layer features \citep{pi0-2025}.

\begin{figure}[!htbp]
	\centering
	\includegraphics[width=0.95\textwidth]{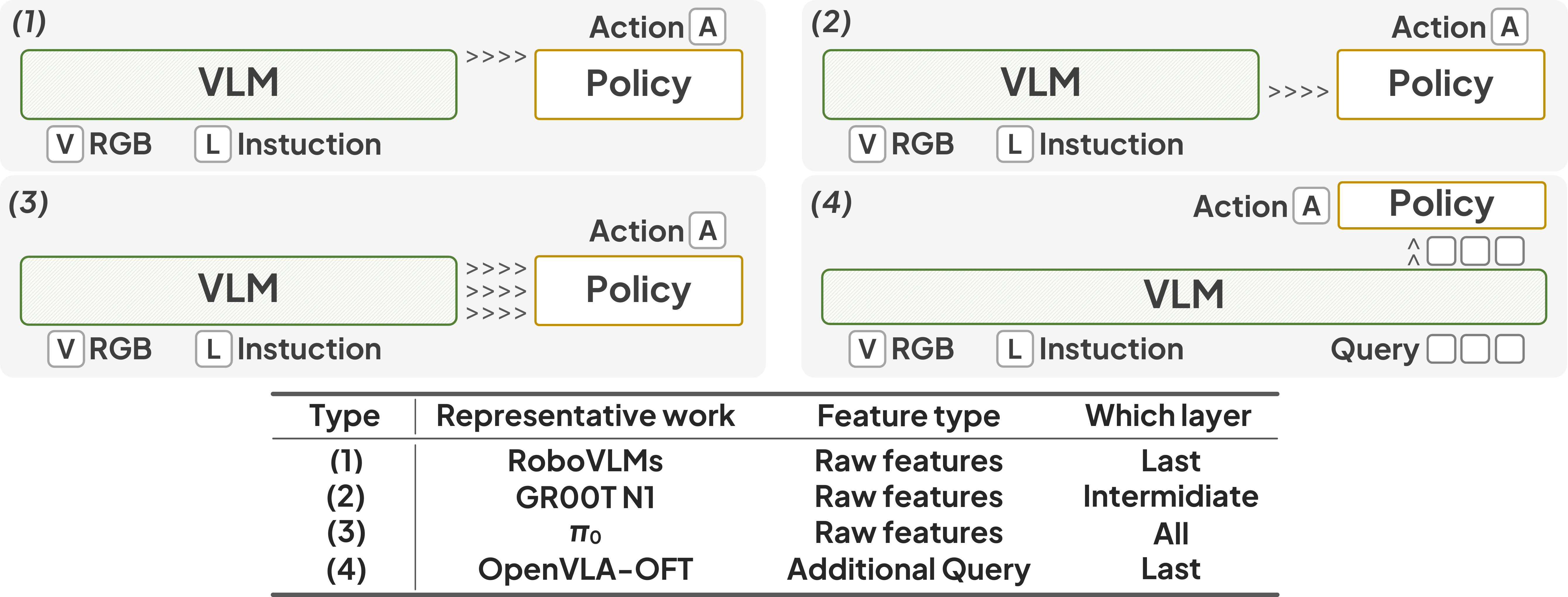}	
	\caption{Existing representative bridge paradigms from VL to A.}\label{Figure_mapping}
\end{figure}

\paragraph{2) Additional Query as Interface.} Furthermore, recent studies \citep{OpenVLA-OFT-2025, OpenHelix-2025} have introduced a novel interface that employs additional queries as bridges between VLMs and Policy, rather than transmitting Raw features. The query is learnable and can incorporate multimodal information, showing superior performance. The existing bridge paradigms are shown in Figure \ref{Figure_mapping}.

%% file: Sections/Methodology.tex
\section{VLA-Adapter Methodology}\label{Section_methodology}

\subsection{Preliminary} \label{Subsection_preliminary}
\begin{figure}[!htbp]
	\centering
	\includegraphics[width=0.97\textwidth]{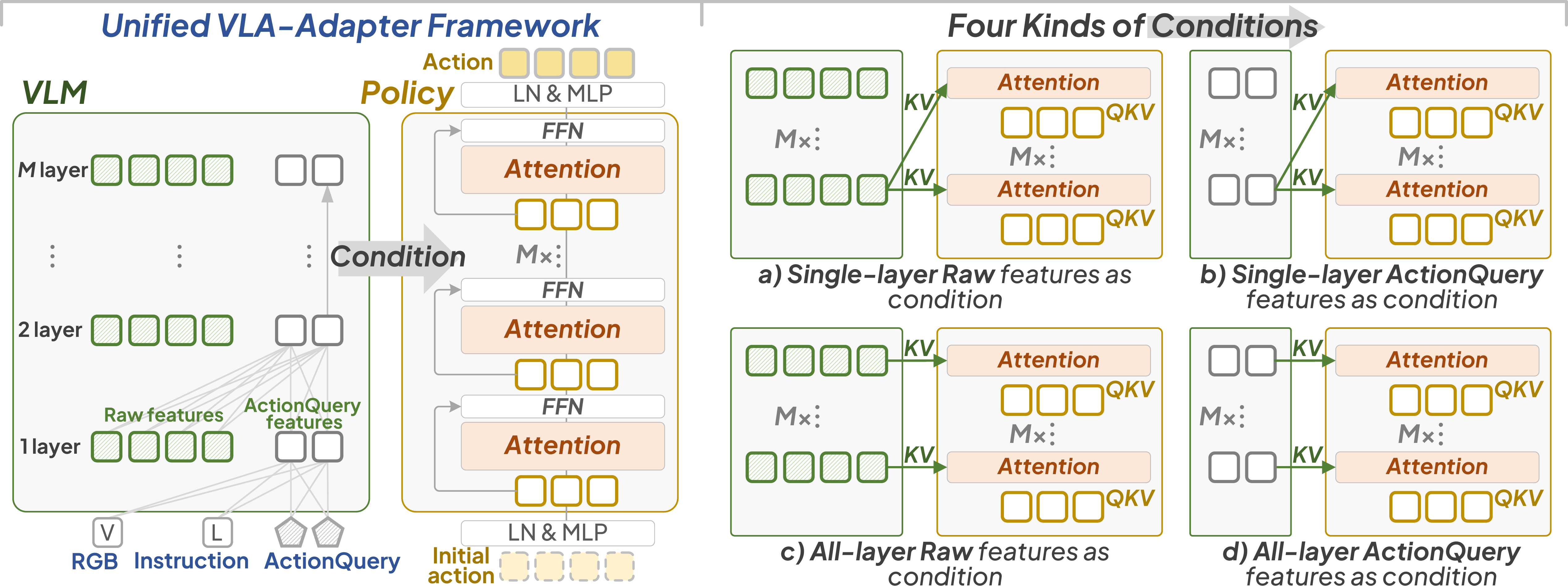}	
	\caption{The proposed VLA framework. The key components are the effective condition exploration and Attention design. ``\textit{Attention}" specifically includes cross attention with conditions and self attention with itself. In the ``\textit{Unified VLA-Adapter Framework}", ``\textit{Attention}" is the Bridge Attention as shown in Section \ref{Subsection_bridge_attention}. Four conditions about ``layer" and ``type" are given on the right.}\label{Figure_framework}
\end{figure}

We present the VLA-Adapter framework, as illustrated in Figure~\ref{Figure_framework}. This VLM follows the Prismatic-VLMs architecture \citep{Prismatic-2024}. It has $M$ layers. At timestep $t$, the input into VLM consists of $\{ \mathcal{X}_t^v, \mathcal{X}_t^g, \mathcal{L}_t, \mathcal{AQ}_t \}$: the 3rd-view image $\mathcal{X}_t^v$, the gripper image $\mathcal{X}_t^g$, the instruction $\mathcal{L}_t$, and additional ActionQuery $\mathcal{AQ}_t$. After inputting $\mathcal{X}_t^v$ and $\mathcal{X}_t^g$, the DINOv2 \citep{DINOv2-2024} and SigLIP \citep{SigLIP-2023} extract vision embeddings. $\mathcal{L}_t$ is tokenized. The outputs are the specified-layer Raw latent $\mathcal{{\cal C}}_t^\mathcal{R}$ and ActionQuery latent $\mathcal{{\cal C}}_t^\mathcal{AQ}$. They serve as the conditions for Policy.

\paragraph{Backbone.} To build a solid basis for research, we perform experiments of VLA-Adapter on different-scale backbones. The backbones select the Prismatic VLM trained on Qwen2.5-0.5B \citep{Qwen2-2024}, the Prismatic VLM trained on LLaMA2-7B \citep{LLaMA2-2023}, and OpenVLA-7B \citep{OpenVLA-2024} pre-trained on robotic data. The benefit gained from increasing backbone scale is limited in VLA-Adapter. The results are shown in Table \ref{Table_effectiveness} of Section \ref{sec41}. Therefore, to ensure efficiency, Qwen2.5-0.5B is our default backbone unless otherwise specified.

\subsection{Which Condition Is Essential for Bridging from VL to A?}\label{Subsection_condition}
Although existing methods have adopted various bridging paradigms from VL to A, their relative effectiveness remains inconclusive. This is mainly due to the differences in the design of the VLM and the Policy. To address this gap, we explore which type of perception information is essential for action generation in the Policy network. In summary, we mainly focus on the following questions:

\begin{description}
	\item[] \thinspace \textit{\textbf{Question 1.1.} Which layer of features within the VLM is more effective for the Policy network? }
	\item[] \thinspace \textit{\textbf{Question 1.2.} Are the ActionQuery features a better choice than the Raw features?}
\end{description}

To ensure compatibility with existing experimental protocols for representative work (e.g., $\pi_{0}$ \citep{pi0-2025}), we let the number of Policy layers be equal to that of VLM. At each layer of Policy, the action latent undergoes cross-attention with conditions and self-attention with itself. This iterative process ultimately yields the action output. Details of the Policy can be seen in Section \ref{Subsection_bridge_attention}.

\paragraph{Experimental Setting.} 

\begin{wrapfigure}{r}{8.8cm}
	\centering
	\vspace{-0.45cm}
	\includegraphics[width=0.6\textwidth]{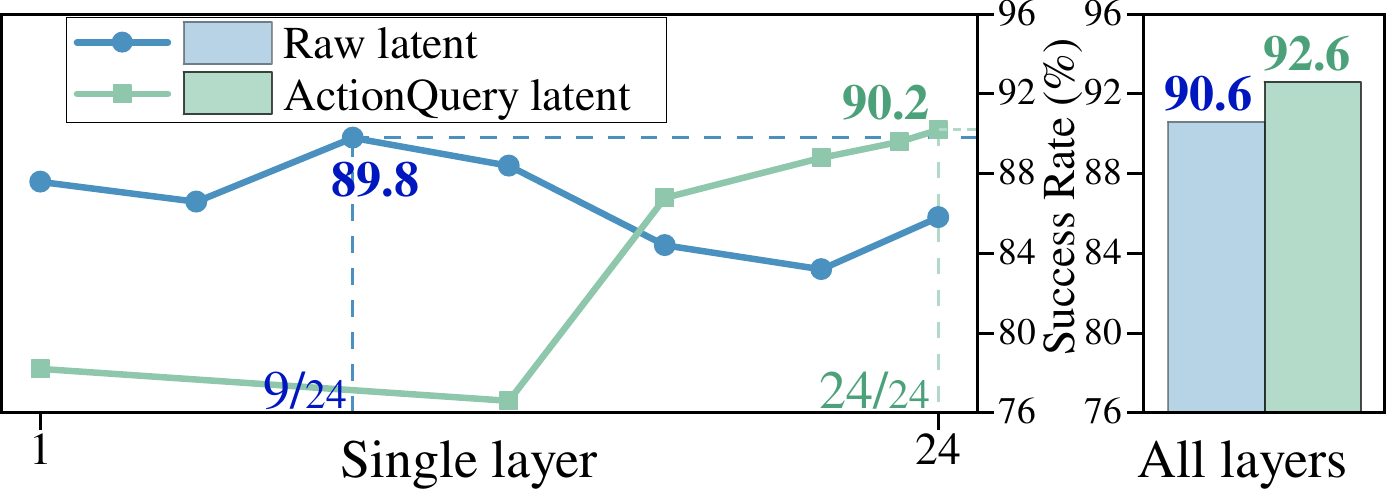}
	\vspace{-0.25cm}
	\caption{Comparison of four conditions in the VLA-Adapter framework on the LIBERO-Long. Blue and Green lines are single-layer $\mathcal{{\cal C}}_t^\mathcal{R}$ and single-layer $\mathcal{{\cal C}}_t^\mathcal{AQ}$, as in Figure \ref{Figure_framework}a) and \ref{Figure_framework}b). Blue and Green columns are all-layer $\mathcal{{\cal C}}_t^\mathcal{R}$ and all-layer $\mathcal{{\cal C}}_t^\mathcal{AQ}$, as in Figure \ref{Figure_framework}c) and \ref{Figure_framework}d). The detailed results are shown in Appendix \ref{AppendixC}. Please note: the number of ActionQuery is 64 here. Its number is variable, similar to MetaQueries \citep{Metaquery-2025} in MLLM research; we will explore it in Section \ref{sec45}.}\label{Figure_condition}
	\vspace{-0.2cm}
\end{wrapfigure}

We evaluate four conditions in our framework. For \textit{\textbf{Question 1.1}}, to evaluate the effectiveness of the individual-layer information, we employ the single-layer latent as the conditions for the all-layer Policy, as shown in Figure \ref{Figure_framework}a) and \ref{Figure_framework}c). To evaluate the effectiveness of all-layer information, we feed each-layer latent into the corresponding-layer Policy, as shown in Figure \ref{Figure_framework}b) and \ref{Figure_framework}d). For \textit{\textbf{Question 1.2}}, to compare the effectiveness of the feature types, we use the $\mathcal{{\cal C}}_t^\mathcal{R}$ or $\mathcal{{\cal C}}_t^\mathcal{AQ}$ as conditions. The comparison on the LIBERO-Long \citep{LIBERO-2023}, which is the long-horizon and complex benchmark, the results are as shown in Figure \ref{Figure_condition}. We give the following key findings.

\begin{description}
	\item[] \thinspace \textit{\textbf{Key Finding 1.} Regarding $\mathcal{{\cal C}}_t^\mathcal{R}$, the middle-layer latent performs better than the deep-layer latent.} Deep-layer $\mathcal{{\cal C}}_t^\mathcal{R}$ is biased towards semantic information and less effective in action generation. The middle-layer $\mathcal{{\cal C}}_t^\mathcal{R}$ effectively integrates image and text information, retains richer multimodal details, and facilitates action generation.
\end{description}

\begin{description}
	\item[] \thinspace \textit{\textbf{Key Finding 2.} Regarding $\mathcal{{\cal C}}_t^\mathcal{AQ}$, deep-layer latent performs better than other-layer latent.} Since ActionQuery is trained from scratch, and deep-layer $\mathcal{{\cal C}}_t^\mathcal{AQ}$ aggregates richer multimodal details and is more effectively promoting action generation than the shallow layers.
\end{description}

\begin{description}
	\item[] \thinspace \textit{\textbf{Key Finding 3.} Multi-layer features perform better.} We observed that using all-layer features generally outperforms a single layer. Not only does it improve performance, but it also saves time on best layer selection during design. This design can be more universal.
\end{description}

\paragraph{Condition Determination.} Does VLA-Adapter rely exclusively $\mathcal{{\cal C}}_t^\mathcal{AQ}$ as conditions? The answer is no. While all-layer $\mathcal{{\cal C}}_t^\mathcal{AQ}$ outperforms $\mathcal{{\cal C}}_t^\mathcal{R}$, middle-layer $\mathcal{{\cal C}}_t^\mathcal{R}$ excels in some hard tasks. Comparison is shown in Table \ref{Table_condition_deter}. So, we aim to enhance performance by using certain knowledge from $\mathcal{{\cal C}}_t^\mathcal{R}$.

\begin{table}[!htbp]
	\centering
    \small
	\caption{Comparison of the $i$th-layer $\mathcal{{\cal C}}_t^\mathcal{R}$ and $\mathcal{{\cal C}}_t^\mathcal{AQ}$ in subtasks of LIBERO-Long.}\label{Table_condition_deter}%
	\begin{tabular}{c|cc|c|ccccccc}
		\toprule[1pt]
		$\mathcal{{\cal C}}_t^\mathcal{R}$ & 9     & 13   &$\mathcal{{\cal C}}_t^\mathcal{AQ}$& 1 & 13    & 17    & 21    & 23    & 24    & All \\
		\midrule[0.5pt]
		Subtask 7& \multicolumn{1}{c}{\textbf{90}} & \multicolumn{1}{c|}{82} & Subtask 7& 76    & 66    & 74    & 70    & 70    & 74    & 76 \\ 
		Subtask 9 &74&\textbf{84}&Subtask 9&78&62&58&72&72&84&78\\
		\bottomrule[1pt]
	\end{tabular}
\end{table}

\subsection{Policy with Bridge Attention}\label{Subsection_bridge_attention}

\begin{figure}[!htbp]
	\centering
	\includegraphics[width=0.97\textwidth]{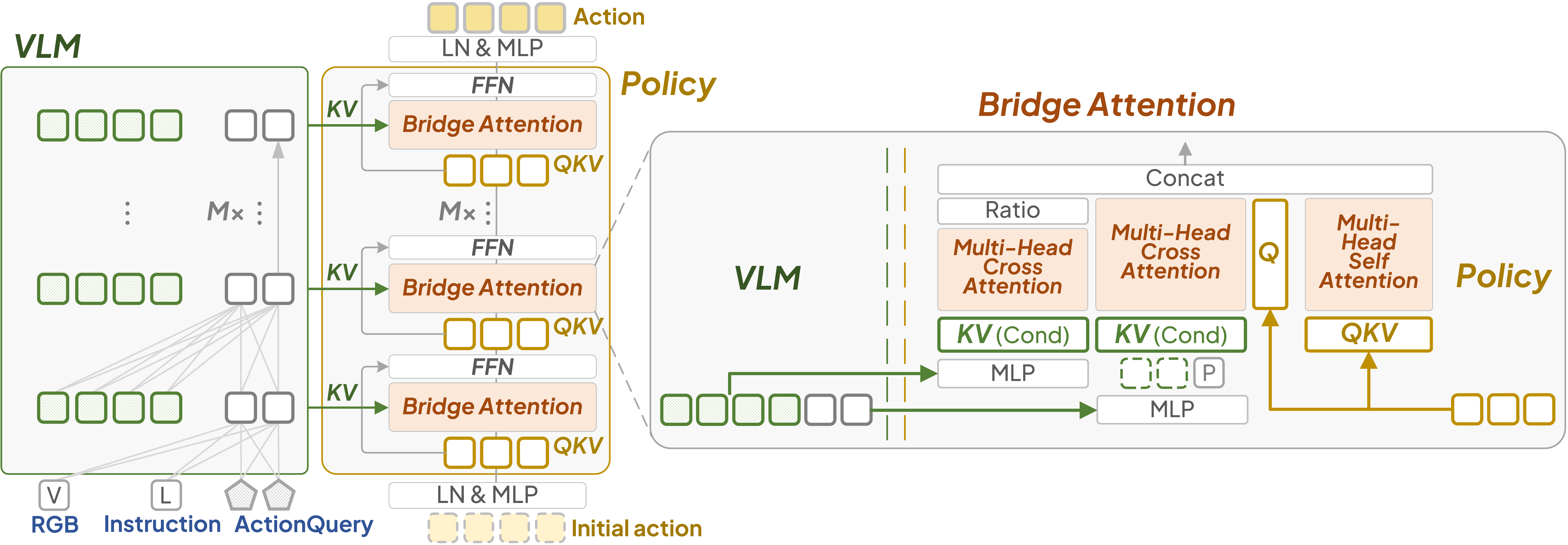}	
	\caption{The Policy with Bridge Attention. The Policy parameters are only 97M when the backbone is Qwen2.5-0.5B. Each-layer $\mathcal{{\cal C}}_t^\mathcal{R}$ and $\mathcal{{\cal C}}_t^\mathcal{AQ}$ are integrated in Bridge Attention with the corresponding-layer action latent. Bridge Attention maps VL to Action to the greatest extent. The degree of $\mathcal{{\cal C}}_t^\mathcal{R}$ injection is learnable, ensuring the performance and stability of training.}\label{Figure_bridge_attention}
\end{figure}

\paragraph{Overall.} For the simplicity of the model, we designed an L1-based Policy network. At $t$-th timestep, the input to Policy includes: $\{ \mathcal{{\cal C}}_t^\mathcal{R}, \mathcal{{\cal C}}_t^\mathcal{AQ}, {\bf A}^{\tau=0}_t , \mathcal{P}_t\} $. 
$\tau$ is the layer of Policy, and it has $\tau\in \mathbb{Z}^+$, $0 \leq \tau \leq M-1$. 
${\bf A}^{0}_t$ is the $H$-step initial action of all zeros, it is processed by LayerNorm (LN) and Multi Layer Perceptron (MLP) to obtain $\widetilde{\bf A}^0_t = \left[\widetilde{\bf a}^0_t, \widetilde{\bf a}^0_{t+1}, \dots, \widetilde{\bf a}^0_{t+H-1} \right]$. $\mathcal{P}_t$ is the proprioceptive state, and it is mapped through a two-layer MLP to obtain the proprio embedding ${\sigma _0}({\mathcal{P}_t})$. The output is the $H$-step action chunk ${\bf A}^{M-1}_t$. Each layer is composed of a Bridge Attention module and a Feed-Forward Network (FFN). The Bridge Attention architecture is shown in Figure \ref{Figure_bridge_attention}.

\paragraph{Bridge Attention.}\label{sec331} 
The proposed Bridge Attention hopes to guide action generation to the greatest extent possible through the conditions $\mathcal{{\cal C}}_t^\mathcal{R}$ and $\mathcal{{\cal C}}_t^\mathcal{AQ}$. Each Bridge Attention consists of two cross attentions and one self attention. In the first cross attention, $\mathcal{{\cal C}}_t^\mathcal{R}$ is processed through an MLP $\sigma_1$ to obtain $K_1,V_1$. The action latent $\widetilde{\bf{A}}^\tau_t$ is used as the $Q_1$, and perform attention to get $\text{CA}_1\left(\widetilde{\bf{A}}^\tau_t,\sigma_1(\mathcal{{\cal C}}_t^\mathcal{R})\right)$. In the second cross attention, $\mathcal{{\cal C}}_t^\mathcal{AQ}$ needs to be concatenated with the ${\sigma _0}({\mathcal{P}_t})$ and passed through an MLP $\sigma_2$ to obtain $K_2,V_2$. $\widetilde{\bf{A}}^\tau_t$ is used as the $Q_2$ to get $\text{CA}_2\left(\widetilde{\bf{A}}^\tau_t,\sigma_2\left[\mathcal{{\cal C}}_t^\mathcal{AQ},{\sigma _0}({\mathcal{P}_t})\right]\right)$. In the self attention, $\widetilde{\bf{A}}^\tau_t$ is as $Q,K,V$, and there is $\text{SA}(\widetilde{\bf{A}}^\tau_t,\widetilde{\bf{A}}^\tau_t)$.

To selectively inject certain $\mathcal{{\cal C}}_t^\mathcal{R}$ into the action space of the Policy, we introduce a learning parameter Ratio $g$ to modulate the influence of $\text{CA}_1\left(\widetilde{\bf{A}}^\tau_t,\sigma_1(\mathcal{{\cal C}}_t^\mathcal{R})\right)$. $g$ is initialized to 0 value, and the $\tanh$ activation function is utilized $\tanh(g)\in[-1,1]$ to prevent extreme values from destabilizing the distribution \citep{LLaMA-Adapter-2024}. And then, the three attentions are concatenated to obtain $\widehat{\bf{A}}_t^\tau$:

\begin{equation}
\widehat{\bf{A}}_t^\tau = 
[\text{CA}_1\left(\widetilde{\bf{A}}^\tau_t,\sigma_1(\mathcal{C}_t^\mathcal{R})\right)\cdot \tanh(g), \text{CA}_2(\widetilde{\bf{A}}^\tau_t,\sigma_2[\mathcal{C}_t^\mathcal{AQ},  \sigma_0({\mathcal{P}_t})]), \text{SA}\left(\widetilde{\bf{A}}^\tau_t, \widetilde{\bf{A}}^\tau_t\right)].
\end{equation}

After Bridge Attention, $\widehat{\bf{A}}_t^\tau$ passes through a residual FFN to obtain $\widetilde{\bf A}^{\tau+1}_t$. Repeating the above process, we finally obtain $\widetilde{\bf A}^{M-1}_t$. The action chunk ${\bf A}^{M-1}_t$ is yielded by an LN and MLP layer. 

Additionally, we also design a DiT-based (Diffusion Transformer \citep{DiT-2023}) Policy. Since the diversity of Policy is not the focus of this paper, we put its details and the brief results in Appendix \ref{Appendix_DiT}. The results show that L1-based performance and inference speed are generally superior to those of the DiT-based approach. Therefore, VLA-Adapter chose the L1 architecture as the Policy.

\subsection{Training}\label{Subection_training}
The training is conducted end-to-end, with the Policy trained from scratch. Given a ground truth action trajectory ${{\bf{A}}_t}$ and action latent ${\bf{A}}_t^{\tau}$. We train VLA-Adapter model $\pi_\theta ( \cdot )$ with the objective:

\begin{equation}
\min_\theta \mathcal{J}(\theta) = \mathbb{E}_{\mathbf{A}_t, \mathcal{C}_t^{\mathcal{R}}, \mathcal{C}_t^{\mathcal{AQ}}, {\sigma _0}({\mathcal{P}_t}),\tau} \Big[\big\| \pi_\theta(\mathbf{A}_t^\tau, \mathcal{C}_t^{\mathcal{R}}, \mathcal{C}_t^{\mathcal{AQ}},{\sigma _0}({\mathcal{P}_t}),\tau) - \mathbf{A}_t \big\|_1\Big].
\end{equation}

For more details of training, please see Appendix \ref{AppendixE1}. 

%% file: Sections/Experiments.tex
\section{Experiments}\label{Section_experiments}

All experiments are run on 4 NVIDIA H100 GPUs. For more details of the hyperparameters, please see Appendix \ref{AppendixE2}. We perform rich experiments to answer the following questions:

\begin{description}
	\item[] \thinspace \textit{\textbf{Question 2.1.} What are the advantages of the VLA-Adapter compared to other bridge paradigms?}
	\item[] \thinspace \textit{\textbf{Question 2.2.} How does VLA-Adapter perform compared to existing methods?}
	\item[] \thinspace \textit{\textbf{Question 2.3.} What else key components in the VLA-Adapter paradigm are worth exploring?}
\end{description}

\paragraph{Experiment Overview.} In Section \ref{sec41}, we use the long-horizon and complex LIBERO-Long \citep{LIBERO-2023}, which typically has a low success rate, to investigate the necessity of VLA-Adapter. From Section \ref{sec42} to Section \ref{sec44}, we use LIBERO \citep{LIBERO-2023} and CALVIN \citep{CALVIN-2022}, which are widely used in VLA, as well as real-world data, to compare the performance comprehensively. In Section \ref{sec45}, we use LIBERO-Long to explore key parts of VLA-Adapter.

\subsection{Necessity of VLA-Adapter}\label{sec41}

\paragraph{Effectiveness of our bridge paradigm.} To validate the effectiveness, we compare three kinds of backbones: · \textit{\textbf{B1}}: The Prismatic VLM \citep{Prismatic-2024} trained on Qwen2.5-0.5B \citep{Qwen2-2024}. · \textit{\textbf{B2}}: The Prismatic VLM trained on LLaMA2-7B \citep{LLaMA2-2023}. The first two are different-scale backbones without pre-training on robotic data. · \textit{\textbf{B3}}: The OpenVLA-7B \citep{OpenVLA-2024} pre-trained on robotic data. We adopted the OpenVLA-OFT bridging paradigm \citep{OpenVLA-OFT-2025} for comparison. It is the existing state-of-the-art level method on major benchmarks, including LIBERO-Long \citep{LIBERO-2023}. The comparison results are shown in Table \ref{Table_effectiveness}.

\begin{table}[!htbp]
	\centering
    \small
    \setlength{\tabcolsep}{1.7mm}
	\caption{Effectiveness comparison with OpenVLA-OFT \citep{OpenVLA-OFT-2025} on the LIBERO-Long \citep{LIBERO-2023}. ``Fine-tuned" is by LoRA fine-tuning \citep{LoRA-2022}. \textbf{Bold} represents the best performance. Please note, comparison with the bridge paradigms of $\pi_0$ \citep{pi0-2025} and GR00T N1 \citep{GR00T-N1-2025} has been included in Section \ref{Section_methodology}, so we will not compare it here.}\label{Table_effectiveness}
	\begin{tabular}{c|cc|cc|cc}
		\toprule[1pt]
		Fine-tuned &  \textit{\textbf{B1}} +OFT   & \textit{\textbf{B1}} +Ours  & \textit{\textbf{B2}} +OFT   & \textit{\textbf{B2}} +Ours  & \textit{\textbf{B3}} +OFT   & \textit{\textbf{B3}} +Ours \\
		\midrule[0.5pt]
		Success Rate (\%) $\uparrow$    & 85.8  & \textbf{95.0} (9.2\% $\uparrow$)  & 87.5  &    \textbf{95.2} (7.7\% $\uparrow$)  & 94.5  & \textbf{95.4} (0.9\% $\uparrow$)\\
		\bottomrule[1pt]
	\end{tabular}%
\end{table}%

Fortunately, VLA-Adapter remains effective when the backbone is frozen. Only the ActionQuery and Policy are trained from scratch. SmolVLA \citep{SmolVLA-2025} is the VLA dedicated to studying frozen VLMs. So, we compare with OpenVLA-OFT and SmolVLA. The results are shown in Table \ref{Table_frozen}. Since the results of GR00T N1 come from \citep{Hume-2025}, it did a full-params tuning, so we will not compare with it here. Based on Tables \ref{Table_effectiveness} and \ref{Table_frozen}, we summarize two conclusions:

\begin{table}[t]
	\centering
    \small
	\caption{Effectiveness comparison when the backbone is frozen. Benchmark is the same as Table \ref{Table_effectiveness}. For a detailed analysis of OpenVLA-OFT \citep{OpenVLA-OFT-2025} does not work, please see Appendix \ref{AppendixG}.}\label{Table_frozen}
	\begin{tabular}{c|ccc}
		\toprule[1pt]
		Frozen & OpenVLA-OFT&SmolVLA&\textbf{VLA-Adapter}\\
		\midrule[0.5pt]
		\multicolumn{1}{c|}{Success Rate (\%) $\uparrow$}  & 0.0 & 77.0 &\textbf{86.4}\\
		\bottomrule[1pt]
	\end{tabular}%
\end{table}%

\begin{description}
	\item[] \thinspace \textit{\textbf{Conclusion 1.} VLA-Adapter improvement is obvious when VLMs without robotic pre-training.}
	\item[] \thinspace \textit{\textbf{Conclusion 2.} Even if the backbone freezes, VLA-Adapter still performs strongly.}
\end{description}

This can be attributed to the fact that, after pre-training on robotic data, the last-layer features are already adapted to the action domain, enabling efficient fine-tuning with a simple MLP. However, when VLMs without pre-training, relying solely on the last-layer latents, are insufficient for effective action mapping. So, adopting the VLA-Adapter becomes crucial to achieve efficient fine-tuning. These insights highlight a \textbf{\textit{key advantage}}: VLA-Adapter facilitates efficient fine-tuning of VLMs without robotic pre-training, achieving performance that surpasses baselines using a tiny backbone.

\paragraph{Efficiency.} VLA-Adapter attains a faster inference speed. The comparison is shown in Table \ref{Table_inference}.

\begin{table}[!htbp]
    \centering
    \small
    \setlength{\tabcolsep}{2mm}
    \caption{Inference efficiency comparison with OpenVLA \citep{OpenVLA-2024} and OpenVLA-OFT \citep{OpenVLA-OFT-2025}. The action chunk is 8 dimensions, consistent with most VLA. ``OpenVLA-OFT (wo $\mathcal{X}_t^g$, $\mathcal{P}$)" is the L1-based version where the input is without the gripper image and proprioceptive state. It is the fastest version of OpenVLA-OFT. Benchmark is the same as Table \ref{Table_effectiveness}.}\label{Table_inference}
	\begin{tabular}{c|ccccc}
		\toprule[1pt]
		Efficiency &OpenVLA&OpenVLA-OFT (wo $\mathcal{X}_t^g$, $\mathcal{P}$) & OpenVLA-OFT&\textbf{VLA-Adapter}\\
		\midrule[0.5pt]
		Throughput (Hz) $\uparrow$&4.2& 109.7 &71.4&\textbf{219.2} \\
		Latency (Sec) $\downarrow$&0.2396&0.0729&0.1120&\textbf{0.0365} \\
		\bottomrule[1pt]
	\end{tabular}%
\end{table}%

\subsection{Overall Performance on Various Tasks}\label{sec42}

\paragraph{Benchmark.}\label{sec421} 
We selected the widely adopted LIBERO benchmark \citep{LIBERO-2023} to evaluate performance across various types of tasks. LIBERO\footnote{\url{https://libero-project.github.io/datasets}} provides multiple suites, including Spatial, Object, Goal, and Long. For detailed settings and examples of LIBERO, please see Appendix \ref{Appendix_LIBERO_benchmark}.

\paragraph{Baselines.}\label{Longtask_baselines} 

We selected recently published, comprehensive, and high-performance VLA works as comparison baselines. They are \textbf{Large}: 1. FlowVLA \citep{FlowVLA-2025}, 2. UnifiedVLA \citep{UnifiedVLA-2025}, 3. OpenVLA \citep{OpenVLA-2024}, 4. OpenVLA-OFT \citep{OpenVLA-OFT-2025}, 5. UniVLA \citep{UniVLA-2025}, 6. CoT-VLA \citep{CoT-VLA-2025}, 7. WorldVLA \citep{WorldVLA-2025}, 8. TraceVLA \citep{TraceVLA-2025}, 9. MolmoAct \citep{MolmoAct-2025}, 10. ThinkAct \citep{ThinkAct-2025}, and 11. PD-VLA \citep{PDVLA-2025}; \textbf{Small}: 12. 4D-VLA \citep{4D-VLA-2025}, 13. SpatialVLA \citep{SpatialVLA-2025}, 14. $\pi_0$ \citep{pi0-2025}, 15. $\pi_0$-FAST \citep{pi0-FAST-2025}, 16. NORA \citep{NORA-2025}, 17. SmolVLA \citep{SmolVLA-2025}, 18. GR00T N1 \citep{GR00T-N1-2025}, and 19. GraspVLA \citep{GraspVLA-2025}; \textbf{Tiny}: 20. Seer \citep{Seer-2025}, 21. VLA-OS \citep{VLA-OS-2025}, and 22. Diffusion Policy \citep{DP-2023}. Their performances are all derived from original references or the reproduction of other published works, ensuring objectivity and accuracy. 

\paragraph{Metrics.}\label{Longtask_metrics1} 
Each subtask is repeated 50$\times$ times to evaluate. We use the commonly used metric ``Success Rate", reported as ranging from 0 to 100, with higher values meaning better performance.

\paragraph{Results.} Comparison on the LIBERO is shown in Table \ref{ComparisonLIBERO}. The results in Table \ref{ComparisonLIBERO} demonstrate that VLA-Adapter, using only a tiny-scale backbone, can achieve performance comparable to OpenVLA-OFT with 14$\times$ larger. It surpasses representative works such as $\pi_0$, SmolVLA, and GR00T N1. In addition, VLA-Adapter has a notable advantage of 29.0\% over VLA-OS with the same-scale backbone on LIBERO-Long. These demonstrate the VLA-Adapter superiority on various tasks.

\begin{table}[t]
    \centering
    \small
    \setlength{\tabcolsep}{2mm}
    \caption{Comparison on the LIBERO benchmark. \textbf{Bold\textsuperscript{*}} is the best performance, \textbf{Bold} is the suboptimal performance, and \underline{\textit{Italics}} is the third best performance. $\dag$ represents that the non-based-VLM baselines. ``Scratch" is the work without pre-training on robotic data. ``Params" is the backbone scale, and its unit is ``\textbf{B}illion". We give the performance on subtasks. It is shown in Table \ref{TableD1} of Appendix \ref{AppendixD}. Recently, we have updated the \textbf{VLA-Adapter-Pro} model. Its Policy architecture is the same as Figure \ref{Figure_bridge_attention}, and we optimized the implementation. For its details, please see Appendix \ref{AppendixI}.}\label{ComparisonLIBERO}
	\begin{tabular}{ccc|cccc|c}
		\toprule[1pt]
		\multicolumn{2}{c}{LIBERO} &Params& \multicolumn{1}{c}{Spatial} & \multicolumn{1}{c}{Object} & \multicolumn{1}{c}{Goal} & \multicolumn{1}{c|}{Long} & \multicolumn{1}{c}{Avg.} \\
		\midrule[0.5pt]
		\multirow{11}{*}{\textit{Large}}&FlowVLA \citep{FlowVLA-2025} \textsubscript{\textit{(ArXiv)}}&8.5&93.2 &95.0 &91.6 &72.6& 88.1\\
        &UnifiedVLA \citep{UnifiedVLA-2025} \textsubscript{\textit{(ArXiv)}}&8.5 & \multicolumn{1}{c}{95.4} & \multicolumn{1}{c}{\underline{\textit{98.8}}} & \multicolumn{1}{c}{93.6} & \multicolumn{1}{c|}{94.0} & \multicolumn{1}{c}{95.5} \\
        &OpenVLA \citep{OpenVLA-2024} \textsubscript{\textit{(CoRL)}}&7 & \multicolumn{1}{c}{84.7} & \multicolumn{1}{c}{88.4} & \multicolumn{1}{c}{79.2} & \multicolumn{1}{c|}{53.7} & \multicolumn{1}{c}{76.5} \\
		&OpenVLA-OFT \citep{OpenVLA-OFT-2025} \textsubscript{\textit{(RSS)}}& 7& \underline{\textit{97.6}} & 98.4 & \multicolumn{1}{c}{\textbf{97.9}} & \multicolumn{1}{c|}{\underline{\textit{94.5}}} & \underline{\textit{97.1}}  \\	
		&UniVLA \citep{UniVLA-2025} \textsubscript{\textit{(RSS)}}&7& 96.5  & 96.8  & 95.6  & \multicolumn{1}{c|}{92.0} & 95.2 \\
		&CoT-VLA \citep{CoT-VLA-2025} \textsubscript{\textit{(CVPR)}}&7 & \multicolumn{1}{c}{87.5} & \multicolumn{1}{c}{91.6} & \multicolumn{1}{c}{87.6} & \multicolumn{1}{c|}{69.0} & \multicolumn{1}{c}{81.1} \\
		&WorldVLA \citep{WorldVLA-2025} \textsubscript{\textit{(ArXiv)}}&7& 87.6  & 96.2  & 83.4  & \multicolumn{1}{c|}{60.0} & 81.8 \\	
		&TraceVLA \citep{TraceVLA-2025} \textsubscript{\textit{(ArXiv)}}&7 & \multicolumn{1}{c}{84.6} & \multicolumn{1}{c}{85.2} & \multicolumn{1}{c}{75.1} & \multicolumn{1}{c|}{54.1} & \multicolumn{1}{c}{74.8} \\	
        &MolmoAct \citep{MolmoAct-2025} \textsubscript{\textit{(ArXiv)}}&7 & \multicolumn{1}{c}{87.0} & \multicolumn{1}{c}{95.4} & \multicolumn{1}{c}{87.6} & \multicolumn{1}{c|}{77.2} & \multicolumn{1}{c}{86.6} \\	
        &ThinkAct \citep{ThinkAct-2025} \textsubscript{\textit{(ArXiv)}}&7 & \multicolumn{1}{c}{88.3} & \multicolumn{1}{c}{91.4} & \multicolumn{1}{c}{87.1} & \multicolumn{1}{c|}{70.9} & \multicolumn{1}{c}{84.4} \\
        & PD-VLA \citep{PDVLA-2025} \textsubscript{\textit{(ArXiv)}} &7 & 95.5& 96.7&94.9&91.7&94.7\\
		\midrule[0.5pt]
		
		\multirow{8}{*}{\textit{Small}}&4D-VLA \citep{4D-VLA-2025} \textsubscript{\textit{(ArXiv)}}&4&88.9&95.2& 90.9& 79.1& 88.6\\
		&SpatialVLA \citep{SpatialVLA-2025} \textsubscript{\textit{(RSS)}}&4& \multicolumn{1}{c}{88.2} & \multicolumn{1}{c}{89.9} & \multicolumn{1}{c}{78.6} & \multicolumn{1}{c|}{55.5} & \multicolumn{1}{c}{78.1} \\
		&$\pi_0$ \citep{pi0-2025} \textsubscript{\textit{(RSS)}}& 3&  \multicolumn{1}{c}{96.8} & \underline{\textit{98.8}} & \multicolumn{1}{c}{95.8} & \multicolumn{1}{c|}{85.2} & \multicolumn{1}{c}{94.2} \\
		&$\pi_0$-FAST \citep{pi0-FAST-2025} \textsubscript{\textit{(RSS)}}& 3 & \multicolumn{1}{c}{96.4} & \multicolumn{1}{c}{96.8} & \multicolumn{1}{c}{88.6} & \multicolumn{1}{c|}{60.2} & \multicolumn{1}{c}{85.5} \\
        &NORA \citep{NORA-2025} \textsubscript{\textit{(ArXiv)}}& 3 & \multicolumn{1}{c}{92.2} & \multicolumn{1}{c}{95.4} & \multicolumn{1}{c}{89.4} & \multicolumn{1}{c|}{74.6} & \multicolumn{1}{c}{87.9} \\
		&SmolVLA \citep{SmolVLA-2025}
		\textsubscript{\textit{(ArXiv)}}& 2.2 & 93.0    & 94.0    & 91.0    & \multicolumn{1}{c|}{77.0} & 88.8 \\	
		&GR00T N1 \citep{GR00T-N1-2025} \textsubscript{\textit{(ArXiv)}}&2 & \multicolumn{1}{c}{94.4} & \multicolumn{1}{c}{97.6} & \multicolumn{1}{c}{93.0} & \multicolumn{1}{c|}{90.6} & 93.9\\
        &GraspVLA \citep{GraspVLA-2025} \textsubscript{\textit{(ArXiv)}}& 1.8 & - & 94.1&91.2&82.0 &89.1\\
		\midrule[0.5pt]
		
		\multirow{5}{*}{\textit{Tiny}}&Seer$^{\dag}$ \citep{Seer-2025} (Scratch)  \textsubscript{\textit{(ICLR)}}&0.57  & \multicolumn{1}{c}{-} & \multicolumn{1}{c}{-} & \multicolumn{1}{c}{-} & \multicolumn{1}{c|}{78.7} & 78.7 \\
		&VLA-OS \citep{VLA-OS-2025} \textsubscript{\textit{(ArXiv)}}&0.5&87.0& 96.5 &92.7 &66.0 &85.6\\
		&Diffusion Policy$^{\dag}$ \citep{DP-2023} \textsubscript{\textit{(RSS)}}& - &  \multicolumn{1}{c}{78.3} & \multicolumn{1}{c}{92.5} & \multicolumn{1}{c}{68.3} & \multicolumn{1}{c|}{50.5} & \multicolumn{1}{c}{72.4} \\	
		
		\rowcolor[rgb]{ .900,  .900,  .900}&\textbf{VLA-Adapter (Ours)}&\textbf{0.5}&\textbf{97.8}&\textbf{99.2}&\underline{\textit{97.2}}&\textbf{95.0}& \textbf{97.3}\\
        \rowcolor[rgb]{ .900,  .900,  .900}&\textbf{VLA-Adapter-Pro (Ours)}&\textbf{0.5}&\textbf{ 99.6\textsuperscript{*}}&\textbf{ 99.6\textsuperscript{*}}&\textbf{ 98.2\textsuperscript{*}}&\textbf{ 96.4\textsuperscript{*}}& \textbf{ 98.5\textsuperscript{*}}\\
		\bottomrule[1pt]
	\end{tabular}%
\end{table}%

\subsection{Performance on Generalization Tasks}\label{sec43}
We used the CALVIN ABC$\to$D \citep{CALVIN-2022} to evaluate the performance on the zero-shot generalization tasks. CALVIN consists of four environments (Env A, B, C, and D)\footnote{\url{http://calvin.cs.uni-freiburg.de/}}
. ``ABC$\to$D" means it trains on Env A, B, and C and evaluates on Env D. VLA needs to execute a preset sequence of 1,000 tasks in sequence. Each task row consists of five subtasks. The model can only proceed to the next subtask after completing the current one. Please see Appendix \ref{AppendixCALVIN} for more settings. 

\paragraph{Baselines.}\label{Generalizationtask_baselines} 

We selected recently published works as baselines. They are \textbf{Large}: 1. UniVLA \citep{UniVLA-2025}, 2. OpenVLA \citep{OpenVLA-2024}, 3. OpenVLA-OFT \citep{OpenVLA-OFT-2025}, 4. VLAS \citep{VLAS-2025}, 5. LCB \citep{LCB-2024}, 6. RoboDual \citep{RoboDual-2024}, 7. OpenHelix \citep{OpenHelix-2025}, and 8. ReconVLA \citep{ReconVLA-2025}; \textbf{Small}: 9. DeeR \citep{DeeR-2024}, 10. RoboFlamingo \citep{RoboFlamingo-2024}, 11. VPP \citep{VPP-2025}, and 12. SuSIE \citep{SuSIE-2024}; \textbf{Tiny}: 13. MoDE \citep{MoDE-2025} and 14. Seer \citep{Seer-2025}. The results of these baselines are based on original references or other published works, ensuring objectivity and correctness. Since the original OpenVLA-OFT paper \citep{OpenVLA-OFT-2025} did not perform experiments on CALVIN ABC$\to$D, we used its source codes to run 150,000 steps and took the best performance.

\paragraph{Metrics.}\label{Generalizationtask_metrics} 
We use the widely used ``Success Rate" (the same in LIBERO \citep{LIBERO-2023}) and ``Avg. len" of completed tasks (the larger the better, with values between 0-5) as metrics.

\paragraph{Results.}\label{Generalizationtask_baselines1} 
Comparison on the CALVIN is shown in Table \ref{ComparisonCALVIN}. The results in Table \ref{ComparisonCALVIN} show that VLA-Adapter has strong generalization ability, and its average length is better than SOTA baselines.

\begin{table}[!htbp]
    \centering
    \small
	\setlength{\tabcolsep}{1.2mm}
	\caption{Comparison on the CALVIN ABC$\to$D benchmark. \textbf{Bold\textsuperscript{*}} is the best performance, \textbf{Bold} is the suboptimal performance, and \underline{\textit{Italics}} is the third best performance. $\dag$ represents that the non-based-VLM method. Recently, we have updated the \textbf{VLA-Adapter-Pro}. Its Policy architecture is the same as Figure \ref{Figure_bridge_attention}, and we optimized the implementation. For its details, please see Appendix \ref{AppendixI}.}\label{ComparisonCALVIN}%
	\begin{tabular}{ccc|ccccc|c}
		\toprule[1pt]
		\multicolumn{2}{c}{\multirow{2}{*}{CALVIN ABC$\to$D}} &\multirow{2}{*}{Params}& \multicolumn{5}{c|}{Task completed in a row $\uparrow$} & \multirow{2}{*}{Avg. len $\uparrow$} \\
		\multicolumn{3}{c|}{} & 1     & 2     & 3     & 4     & 5     &\\
		\midrule[0.5pt]
		\multirow{8}{*}{\textit{Large}}&UniVLA \citep{UniVLA-2025} \textsubscript{\textit{(RSS)}} &7& 95.5  & 85.8  & 75.4  & 66.9  & 56.5  & 3.80\\
		&OpenVLA \citep{OpenVLA-2024} \textsubscript{\textit{(CoRL)}} &7& 91.3  & 77.8  & 62.0    & 52.1  & 43.5  & 3.27 \\	 
        & OpenVLA-OFT \citep{OpenVLA-OFT-2025} \textsubscript{\textit{(RSS)}}&7&96.3&89.1&82.4&75.8&66.5&4.10\\
		&VLAS \citep{VLAS-2025} \textsubscript{\textit{(ICLR)}}& 7 &87.2 &64.2& 40.9 &28.1 &19.6& 2.40\\
		&LCB \citep{LCB-2024} \textsubscript{\textit{(IROS)}} &7 & 73.6 & 50.2 & 28.5&16.0 & 9.9 & 1.78\\
		&RoboDual \citep{RoboDual-2024} \textsubscript{\textit{(ArXiv)}} &7& 94.4  & 82.7  & 72.1  & 62.4  & 54.4  & 3.66 \\
		&OpenHelix \citep{OpenHelix-2025} \textsubscript{\textit{(ArXiv)}} &7& \underline{\textit{97.1}} & 91.4 & 82.8 & 72.6 & 64.1 & 4.08 \\
        & ReconVLA \citep{ReconVLA-2025} \textsubscript{\textit{(ArXiv)}}&7&95.6 &87.6& 76.9& 69.3 &64.1& 3.95\\
        
		\midrule[0.5pt]
		\multirow{4}{*}{\textit{Small}}&DeeR \citep{DeeR-2024} \textsubscript{\textit{(NeurIPS)}} & 3&86.2  & 70.1  & 51.8  & 41.5  & 30.4  & 2.82 \\	
		&RoboFlamingo \citep{RoboFlamingo-2024} \textsubscript{\textit{(ICLR)}} &3& 82.4  & 61.9  & 46.6  & 33.1  & 23.5  & 2.48 \\
        &VPP$^{\dag}$ \citep{VPP-2025} \textsubscript{\textit{(ICML)}} &1.5& 95.7  & 91.2  & \underline{\textit{86.3}} & \underline{\textit{81.0}} & \underline{\textit{75.0}} & \underline{\textit{4.33}} \\
		&SuSIE \citep{SuSIE-2024} \textsubscript{\textit{(ICLR)}} &1.3& 87.0    & 69.0    & 49.0    & 38.0    & 26.0    & 2.69 \\

		\midrule[0.5pt]
		\multirow{5}{*}{\textit{Tiny}}&Seer\textsubscript{\textit{Large}}$^{\dag}$  \citep{Seer-2025} \textsubscript{\textit{(ICLR)}} &0.57& 96.3 & \underline{\textit{91.6}} & 86.1  & 80.3  & 74.0    & 4.28 \\
        &MoDE$^{\dag}$ \citep{MoDE-2025} \textsubscript{\textit{(ICLR)}} &0.44& 96.2  & 88.9  & 81.1  & 71.8  & 63.5  & 4.01 \\
		&Seer$^{\dag}$ \citep{Seer-2025} \textsubscript{\textit{(ICLR)}} &0.32& 94.4  & 87.2  & 79.9  & 72.2  & 64.3  & 3.98 \\	
		\rowcolor[rgb]{ .900,  .900,  .900}&\textbf{VLA-Adapter (Ours) }&\textbf{0.5}&\textbf{ 99.1\textsuperscript{*}} &\textbf{94.6}&\textbf{88.8}&\textbf{82.8}&\textbf{76.5}&\textbf{4.42} \\
        \rowcolor[rgb]{ .900,  .900,  .900}&\textbf{VLA-Adapter-Pro (Ours) }&\textbf{0.5}&\textbf{98.5} &\textbf{ 95.0\textsuperscript{*}}&\textbf{ 90.5\textsuperscript{*}}&\textbf{ 85.3\textsuperscript{*}}&\textbf{ 80.0\textsuperscript{*}}&\textbf{ 4.50\textsuperscript{*}} \\	
		\bottomrule[1pt]
	\end{tabular}%
\end{table}%

\subsection{Performance on Real-World Tasks}\label{sec44}

\begin{figure}[!htbp]
	\centering
	\includegraphics[width=0.95\textwidth]{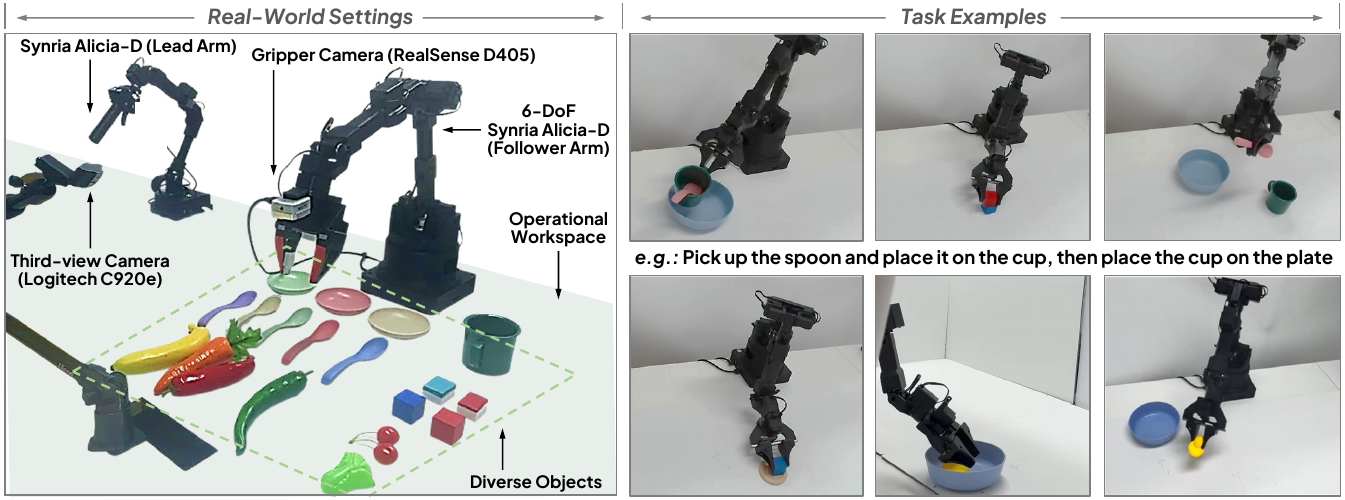}	
	\caption{Real-world system Synria Alicia-D and the task examples.}\label{Figure_realworld}
\end{figure}

\paragraph{Experimental settings.}\label{sec441}
We use a robotic system to perform real-world tasks. A 6-DOF Synria Alicia-D equipped with a 1-DOF gripper is employed, and it uses Logitech C920e and RealSense D405 cameras to capture the third-view and gripper images. The real-world robotic system is shown in Figure \ref{Figure_realworld}. 
We evaluate the VLA-Adapter method across four experimental categories:

\begin{description}
    \item[] \thinspace \textit{\textbf{1)} Simple pick-and-place tasks with objects spanning diverse materials and geometries.}
	\item[] \thinspace \textit{\textbf{2)} CALVIN-inspired challenging task II: lateral block relocation (e.g. ``Move $<$obj$>$ left/right").}
    \item[] \thinspace \textit{\textbf{3)} CALVIN-inspired challenging manipulation task I: ``Block stacking".}
    \item[] \thinspace \textit{\textbf{4)} LIBERO-inspired complex and long-horizon task: (e.g. ``Pick up the spoon and place it on the cup, then place the cup on the plate").}  
\end{description}

To strengthen evaluation rigor and assess generalization performance, we randomize the object positions at test time to induce distribution shift and increase task difficulty.

\paragraph{Baselines.}\label{Real-world baselines} 
ACT \citep{ACT-2023} and OFT-style variant \citep{OpenVLA-OFT-2025} are as baselines.

\begin{figure}[!htbp]
	\centering
	\includegraphics[width=0.87\textwidth]{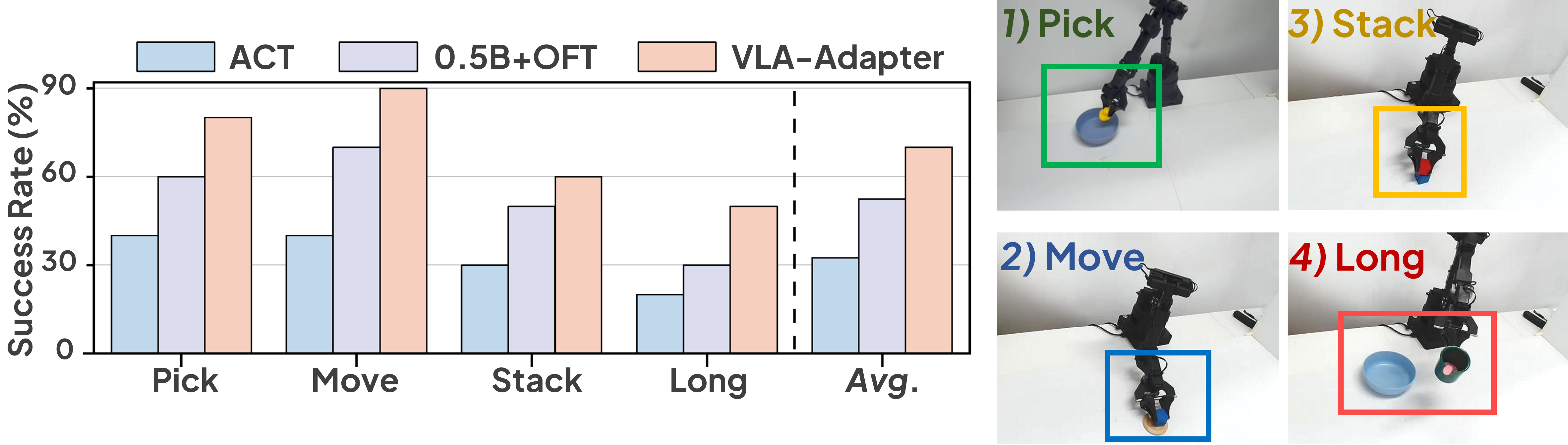}
	\caption{Comparison on real-world tasks.}\label{Figure_realworld_results}
\end{figure}

\paragraph{Results.}
The comparison results are shown in Figure \ref{Figure_realworld_results}. Each result is obtained by averaging the results of 10 executions.
Experimental results show that VLA-Adapter has better generalization capabilities in various scenarios. Therefore, VLA-Adapter greatly lowers the barrier to adopting VLA in practical applications. More real-world experiments are detailed in Appendix \ref{AppendixF}.

\subsection{Ablation Experiments}\label{sec45}

We explore three key components in the VLA-Adapter: 1. Number of ActionQuery, 2. Condition type, and 3. Injection degree for Policy. The benchmark is LIBERO-Long \citep{LIBERO-2023}.

\begin{figure}[!htbp]
	\centering
	\includegraphics[width=0.63\textwidth]{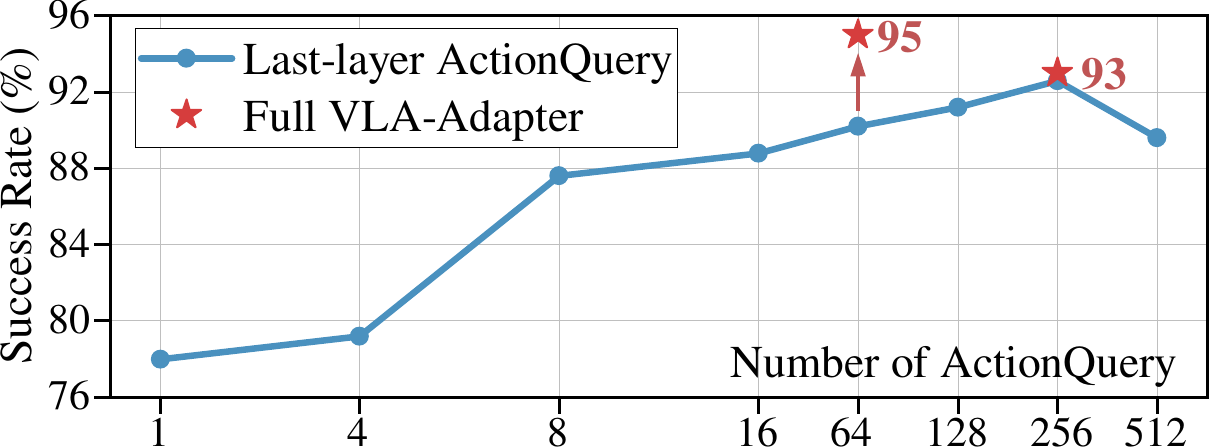}
	\caption{Comparison of the different numbers of ActionQuery. The blue line shows the result of using only the last-layer ActionQuery. The red star shows the result of the full VLA-Adapter.}\label{Figure_ActionQuery}
\end{figure}

\paragraph{Number of ActionQuery.}
In our paradigm, the number of ActionQuery is not fixed. To explore the impact of this number on performance, we conducted the following experiments by varying the number of ActionQuery to 1, 4, 8, 16, 64, 128, 256, and 512. The results are shown in Figure \ref{Figure_ActionQuery}. Thus, using too few ActionQuery tokens weakens multimodal aggregation and makes it challenging to condition the Policy. Conversely, employing too many ActionQuery tokens introduces redundancy, interfering with the performance. Therefore, we selected 64 ActionQuery tokens. This number provides the optimal balance between performance and efficiency.

\paragraph{Condition Type.}
In Section \ref{Section_methodology}, we analyzed the overall effects of different conditions on action generation. Here, we present the complete comparison results based on the four classic paradigms in Section \ref{Section_related_work}, as shown in Table \ref{Table_condition_type}. This result demonstrates that using both all-layer Raw and ActionQuery achieves superior performance, indirectly validating the superiority of our bridge paradigm.

\begin{table}[!htbp]
	\centering
	\caption{Comparison with different condition types. The style can be summarized as representative works in Figure \ref{Figure_mapping} of Section \ref{Section_introduction}. ``\textit{N/A}" represents no such method. ``\textbf{Bold}" is the best performance.}\label{Table_condition_type}%
	\begin{tabular}{c|cccc}
		\toprule[1pt]
		Layer  &  Raw   & ActionQuery &Style & \multicolumn{1}{c}{SR $\uparrow$} \\
		\midrule[0.5pt]
		\multicolumn{1}{c|}{\multirow{2}{*}{Last}} & \ding{51}     & \ding{55}      & RoboVLMs \citep{RoboVLMs-2024} & 85.8 \\
		&  \ding{55}     & \ding{51}      & OpenVLA-OFT \citep{OpenVLA-OFT-2025}&90.2 \\
		\midrule[0.5pt] 
		\multicolumn{1}{c|}{\multirow{1}{*}{Intermidiate}} & \ding{51}      & \ding{55}     & GR00T N1 \citep{GR00T-N1-2025} & 88.4 \\
		\midrule[0.5pt] 
		\multicolumn{1}{c|}{\multirow{3}{*}{All}} & \ding{51}      & \ding{55}     &$\pi_0$ \citep{pi0-2025}  &  90.6 \\
		& \ding{55}      & \ding{51}     & \textit{N/A}& 92.6 \\
		\rowcolor[rgb]{ .900,  .900,  .900}& \ding{51}      & \ding{51}     & \textbf{VLA-Adapter (Ours)} & \textbf{95.0} \\
		\bottomrule[1pt]
	\end{tabular}%
\end{table}%

\paragraph{Injection Degree for Policy.}\label{sec453}
In the Bridge Attention, we use learnable parameters to control the injection degree of Raw features $\mathcal{{\cal C}}_t^\mathcal{R}$ and set the injection degree of ActionQuery features $\mathcal{{\cal C}}_t^\mathcal{AQ}$ to 1. Here, we explore other injection degrees, and the comparison results are shown in Table \ref{Table_injection}. Two conclusions can be drawn from the results in Table \ref{Table_injection}:
\textit{From 1) and 2)}, the performance of $\mathcal{{\cal C}}_t^\mathcal{R}$ is inferior to $\mathcal{{\cal C}}_t^\mathcal{AQ}$, so $\mathcal{{\cal C}}_t^\mathcal{R}$ should inject some effective information into Policy through learning.
\textit{From 1) and 4)}, $\mathcal{{\cal C}}_t^\mathcal{AQ}$ aggregates multimodal information, which is beneficial for action generation; it needs to be injected fully into Policy. This result confirms that the Bridge Attention is effective.

\begin{table}[htbp]
	\centering
	\caption{Ablation of other injection degrees.}\label{Table_injection}%
	\begin{tabular}{c|ccc}
		\toprule[1pt]
		&Raw   & ActionQuery    & Success Rate (\%) \\
		\midrule[0.5pt]
		\textit{1)} (\textbf{VLA-Adapter})&$\tanh(g)$  & 1 & \textbf{95.0} \\
		\textit{2)}&1 & 1 & 91.4 \\
		\textit{3)}&1 & $\tanh(g)$  & 91.0 \\
		\textit{4)}&$\tanh(g)$  & $\tanh(g)$  & 92.6 \\
		\bottomrule[1pt]
	\end{tabular}%
\end{table}%

%% file: Sections/Conclusion.tex
\section{Conclusion}\label{sec5}

We propose VLA-Adapter, a novel and efficient bridging paradigm for VLA. By leveraging Raw and ActionQuery latent, this method effectively transfers multimodal knowledge to the Policy to generate action.
Experiments show that VLA-Adapter achieves SOTA performance using a tiny-scale backbone. Even when the VLM is frozen, it has strong performance. 
In addition, our method has low VRAM usage and high inference speed.
These results suggest that VLA-Adapter alleviates VLA's reliance on large-scale VLMs and huge training costs, lowering the barrier to deploying VLA.

Ultimately, we hope the VLA-Adapter method and key findings of this study can provide a solid basis for future research in the VLA and inspire the development of more advanced VLA methods!

%% file: Sections/Limitation.tex
\section{Limitations}\label{sec6}

While VLA-Adapter achieves lightweight and excellent performance, it also has some limitations. First, because VLA-Adapter is not pre-trained on a large amount of embodied data and the scale is tiny, its generalization in real-world systems needs to be improved. Secondly, the quality of the actions generated by the Policy networks depends on the conditions provided by the VLM and how they are used. Therefore, future work can further explore these conditions to improve its representation and ensure its efficient use. Finally, the fundamental training process of the VLA-Adapter is still relatively simple, and the complex processes, such as reinforcement learning, can be explored.

%% file: Sections/Acknowledgments.tex
\section*{Acknowledgments}
This work was supported in part by the National Natural Science Foundation of China under Grant U21B2020, and the BUPT Excellent Ph.D. Students Foundation under Grant CX20241055.

%% file: Sections/Appendix.tex
\section*{\centering \textit{\textbf{Appendix of VLA-Adapter}}}

\section{Setup Details of LIBERO Simulation Benchmarks}\label{Appendix_LIBERO_benchmark}

\renewcommand{\thefigure}{A\arabic{figure}}
\renewcommand{\theHfigure}{A\arabic{figure}}
\setcounter{figure}{0}

\renewcommand{\thetable}{A\arabic{table}}
\renewcommand{\theHtable}{A\arabic{table}}
\setcounter{table}{0}

\begin{figure}[!htbp]
	\centering
	\includegraphics[width=0.99\textwidth]{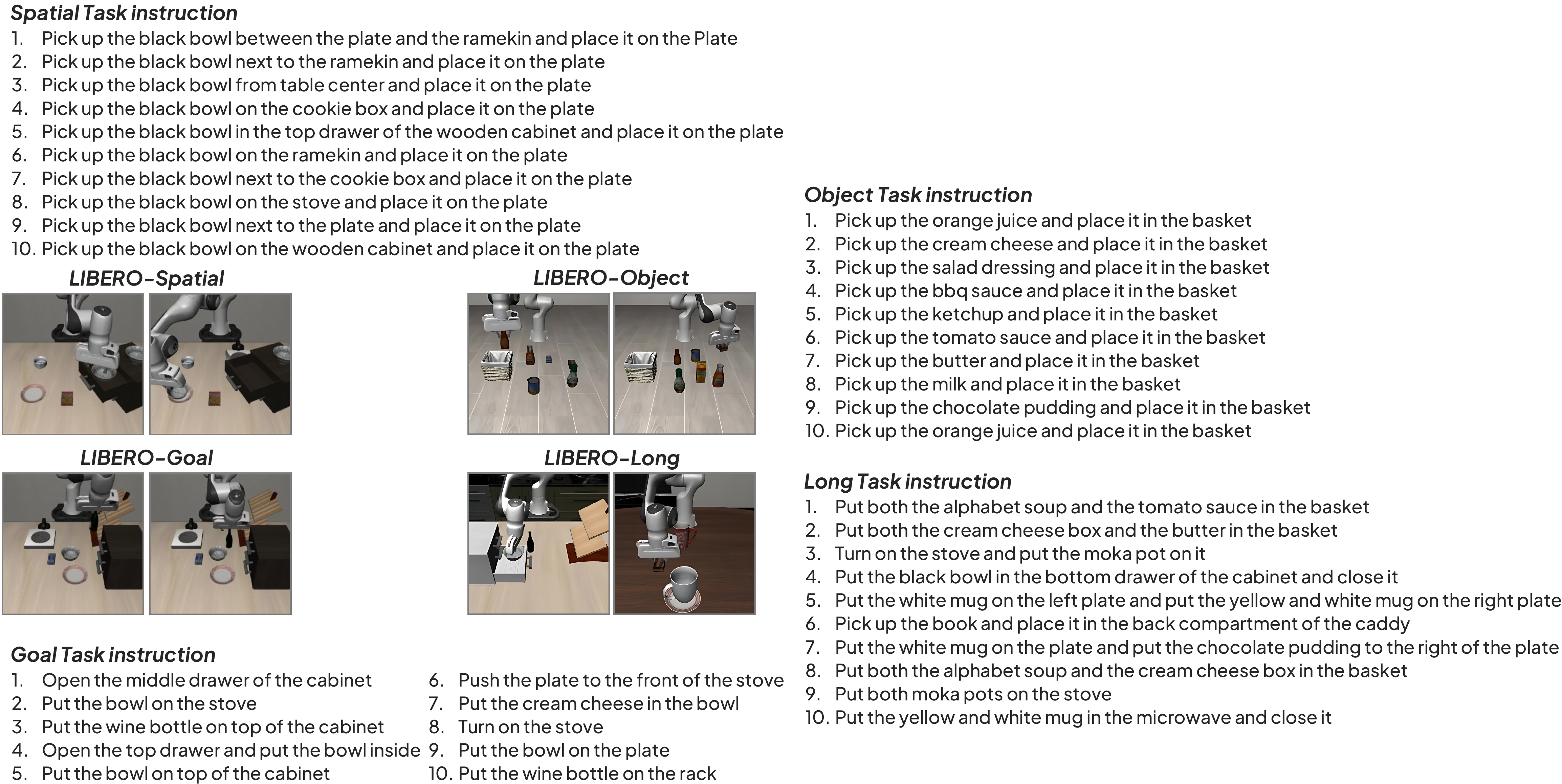}
	\caption{The examples and the task instructions on the LIBERO benchmark.}\label{Figure_LIBERO}
\end{figure}

The LIBERO benchmark \citep{LIBERO-2023} comprises four distinct task suites: LIBERO-Spatial, LIBERO-Object, LIBERO-Goal, and LIBERO-100. The first three suites each contain 10 tasks, and LIBERO-100 contains 90 short-term tasks (LIBERO-90) and 10 long-horizon tasks (LIBERO-Long). The strategy for each task depends solely on the current instructions provided. Each task is repeated multiple times (50 repetitions in this paper) to obtain the average success rate for each subtask. The examples and the instructions in the LIBERO benchmark are shown in Figure \ref{Figure_LIBERO}.

In the LIBERO benchmark, we use third-person images (resolution 224$\times$224$\times$3, RGB) and wrist images (resolution 224$\times$224$\times$3, RGB) as visual input. The task instruction is first constructed as a prompt in a specific format: ``\texttt{In: What action should the robot take to {instruction.lower()}?\textbackslash nOut:}", and then input into the VLM module together with the image information. Its output is a 7-dimensional action vector, which is used to control the 7-DOF Franka Emika Panda simulated robot arm to perform the corresponding action sequence.

\vspace{3ex}

\section{DiT-Based Policy Network}\label{Appendix_DiT}
\renewcommand{\thefigure}{B\arabic{figure}}
\renewcommand{\theHfigure}{B\arabic{figure}}
\setcounter{figure}{0}

\renewcommand{\thetable}{B\arabic{table}}
\renewcommand{\theHtable}{B\arabic{table}}
\setcounter{table}{0}

\renewcommand{\theequation}{B-\arabic{equation}}
\renewcommand{\theHequation}{B-\arabic{equation}}
\setcounter{equation}{0}

\subsection{Overall Architecture}\label{Appendix_DiT_1} 

\begin{figure}[!htbp]
	\centering
	\includegraphics[width=0.45\textwidth]{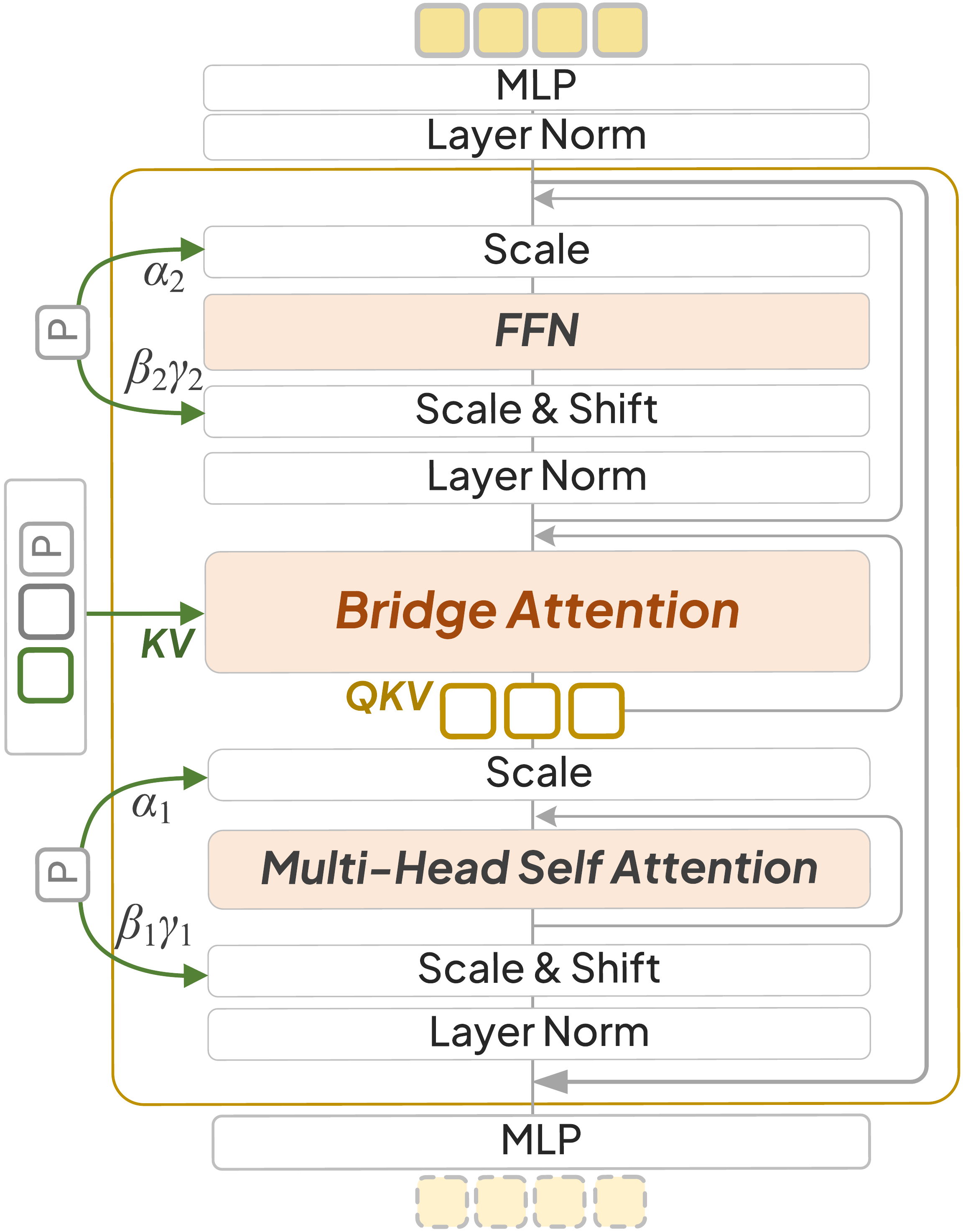}	
	\caption{The DiT-based policy network.}\label{Figure_DiT}
\end{figure}

This architecture is shown in Figure \ref{Figure_DiT}. It consists of $\tau$-DiT blocks, $1\leq\tau\leq M$. It has the same number of layers as VLM. Each DiT block consists of three components: conditional modulation, conditional attention, and a conditional feedforward network.
At timestep $t$, the input chunk to the first-DiT block, ${\bf A}^1_t$, is a noisy action sequence. Since the input contains random noise, to facilitate the transition from ``pure noise $\to$ fine-grained prediction", we adopt the AdaLN-Zero layer \citep{DiT-2023} to modulate the activation amplitude at each layer. The AdaLN-Zero consists of LayerNorm, modulation, and gated residual. Specifically, $C^{\mathcal{M}}_t$ will be obtained by $\mathcal{P}_t$ and $\mathcal{C}_t^{\mathcal{R}}$, i.e., $C^{\mathcal{M}}_t = \sigma'_1(\mathcal{C}_t^{\mathcal{R}}) + \sigma_0(\mathcal{P}_t)$. It is used to generate ``Scale" and ``Shift" vectors, which guide the activation direction of intermediate features and inject automatic modulation amplitude via gated residual control. After modulation, $\widetilde{\bf A}^1_t = \left[ \widetilde{\bf a}_t^1, \widetilde{\bf a}_{t+1}^1, \dots, \widetilde{\bf a}_{t+H-1}^1 \right]$ is obtained:

\begin{equation}\label{AppedixEq4}
\small
\widetilde {\bf A}_t^1  = {\bf{A}}_t^1  + {\alpha _\tau} \odot \varepsilon ({\gamma _\tau }{\mathop{\rm LN}\nolimits} ({\bf{A}}_t^1) + {\beta _\tau })= {\bf{A}}_t^1  + {\sigma' _2}(\mathcal{C}_t^\mathcal{M}) \odot \varepsilon ({\sigma'_1}^{(2)}(\mathcal{C}_t^\mathcal{M}){\mathop{\rm LN}\nolimits} ({\bf{A}}_t^1) + {\sigma'_1}^{(1)}(\mathcal{C}_t^\mathcal{M})), \ [{\beta _\tau };{\gamma _\tau }] = {\sigma' _1}(\mathcal{C}_t^\mathcal{M}),
\end{equation}

\noindent where, ${\beta_\tau}$ and ${\gamma_\tau}$ are scaling factors and offset factors, which are dynamically generated by $\mathcal{C}_t^\mathcal{M}$ through a projection new with the SiLU. ${\alpha_\tau}$ is the gated residual coefficient, used to adjust the injection amplitude. It is dynamically generated by $\mathcal{C}_t^\mathcal{M}$ through ${\sigma_0}$ with the SiLU, and has ${\alpha _\tau} = {\sigma' _3}(\mathcal{C}_t^\mathcal{M})$. $\odot$ is element-wise multiplication, and $\varepsilon(\cdot)$ is self-attention and projection modules.

After conditional modulation, $\widetilde {\bf{A}}_t^1$ is used as the $QKV$ vector, and $\mathcal{C}_t^\mathcal{R},\mathcal{C}_t^\mathcal{AQ}$ are used as the $KV$ vectors for Bridge Attention. The details of Bridge Attention are shown in Section \ref{Subsection_bridge_attention}. And then, the attention latent $\widehat{\bf{A}}^1_t$ will be obtained. $\widehat{\bf{A}}^1_t$ is input into the conditional feedforward network. The first-DiT block output ${\bf A}^2_t$ is obtained. After passing through $M$ DiT blocks, we get the $\widehat{\bf A}^{M}_t$, which is passed by a LayerNorm and MLP layer to generate the current action chunk ${\bf A}^M_t$.

\subsection{Training of DiT-Based Policy}\label{Appendix_DiT_2}
This Policy is also trained from scratch. Given a ground truth action trajectory ${{\bf{A}}_t}$, a noisy action in DiT-based Policy ${\bf{A}}_t^{\tau} = \sqrt {{{\overline \alpha  }_{\tau} }} {{\bf{A}}_{\rm{t}}} + \sqrt {1 - {{\overline \alpha  }_{\tau} }} \epsilon$, where, $\sqrt {{{\overline \alpha  }_{\tau} }} $ is the cumulative product of noise coefficients, and has $\sqrt {{{\overline \alpha  }_{\tau} }}  = \prod\nolimits_{i = 1}^T {{\alpha _i}}  = \prod\nolimits_{i = 1}^T {(1 - {\beta _i})} $, $\beta_i$ is the variances used at each step, and $\epsilon \sim \mathcal{N}(0,{\bf{I}})$ is Gaussian noise. We train DiT-based model $\pi_\theta ( \cdot )$ with the training objectives:

\begin{equation}
\mathop {\min }\limits_\theta \mathcal{J}(\theta ) = {\mathbb{E}_{{{\bf{A}}_t},\epsilon \sim \mathcal{N}(0,{\bf{I}}),\mathcal{{\cal C}}_t^\mathcal{AQ},{\sigma _1}({\mathcal{P}_t}),\tau}}\Big[\big\|{\pi _\theta } \big(\sqrt {{{\overline \alpha  }_{\tau} }} {{\bf{A}}_t}+ \sqrt {1 - {{\overline \alpha  }_{\tau} }} \mathcal{},\mathcal{C}_t^\mathcal{AQ},{\sigma _1}({\mathcal{P}_t}),\tau\big) - \epsilon\big\|_2^2\Big].
\end{equation}

\subsection{Brief Comparison with L1-based Policy}\label{Appendix_DiT_3}
As exploring the Policy architecture is not the primary focus of this paper, we briefly compare the performance of the L1-based and DiT-based Policy networks on the LIBERO-Long benchmark.

\begin{table}[!htbp]
	\centering
	\small
	\caption{Comparison of the L1-based and DiT-based Policy networks of VLA-Adapter. \textbf{Bold} represents the best results. For detailed task instructions, please see Figure \ref{Figure_LIBERO} in Appendix \ref{Appendix_LIBERO_benchmark}.}\label{Table_DiT}%
	\begin{tabular}{c|cccccccccc|c}
		\toprule[1pt]
		\textbf{Task instructions} &1&2&3&4&5&6&7&8&9&10& \textbf{Avg. $\uparrow$}\\
		\midrule[0.5pt]
		\multicolumn{1}{c|}{L1-based} &\multicolumn{1}{c}{96.0}&\multicolumn{1}{c}{\textbf{96.0}}& \multicolumn{1}{c}{\textbf{100.0}} &\multicolumn{1}{c}{\textbf{98.0}}&\multicolumn{1}{c}{\textbf{100.0}}& \multicolumn{1}{c}{\textbf{100.0}}& \multicolumn{1}{c}{\textbf{84.0}}& \multicolumn{1}{c}{96.0}& \multicolumn{1}{c}{\textbf{84.0}}& \multicolumn{1}{c|}{\textbf{96.0}}&\multicolumn{1}{c}{\textbf{95.0}} \\
        
		\multicolumn{1}{c|}{DiT-based} & \multicolumn{1}{c}{96.0}&\multicolumn{1}{c}{92.0}&\multicolumn{1}{c}{98.0}&\multicolumn{1}{c}{96.0}&\multicolumn{1}{c}{90.0}& \multicolumn{1}{c}{98.0} &\multicolumn{1}{c}{82.0}& \multicolumn{1}{c}{\textbf{100.0}} &\multicolumn{1}{c}{74.0}& \multicolumn{1}{c|}{90.0}&\multicolumn{1}{c}{91.6} \\
		\bottomrule[1pt]
	\end{tabular}%
\end{table}%

Results presented in Table \ref{Table_DiT} indicate that the L1-based Policy achieves superior performance compared to the DiT-based Policy. This phenomenon coincides with the conclusions in OpenVLA-OFT \citep{OpenVLA-OFT-2025}: Diffusion-type Policy performs better during pre-training, but L1-based Policy outperforms Diffusion-type Policy during fine-tuning because their actions are less redundant. Furthermore, consistent with findings from OpenVLA-OFT, the L1-based Policy achieves higher throughput compared to the Diffusion-type Policy. Therefore, we chose the L1-based Policy.

\vspace{3ex}

\section{Detailed Comparison Results of Different Coditions}\label{AppendixC}
\renewcommand{\thefigure}{C\arabic{figure}}
\renewcommand{\theHfigure}{C\arabic{figure}}
\setcounter{figure}{0}

\renewcommand{\thetable}{C\arabic{table}}
\renewcommand{\theHtable}{C\arabic{table}}
\setcounter{table}{0}

In this section, we give the specific performance of the ten subtasks on the LIBERO-Long benchmark in Section \ref{Section_methodology} to explore different conditions. The specific results are shown in Tables \ref{TableC1} and \ref{TableC2}.

\begin{table}[!htbp]
	\centering
    \small
	\caption{The specific performance of different layers of Raw features on 10 subtasks. For detailed task instructions, please see Figure \ref{Figure_LIBERO} in Appendix \ref{Appendix_LIBERO_benchmark}. \textbf{Bold} represents the best performance of the average success rate. \underline{\textit{Italics}\textsuperscript{*}} represents the suboptimal performance of the average success rate.}\label{TableC1}%
	\begin{tabular}{cc|cccccccccc|c}
		\toprule[1pt]
		\multicolumn{2}{c|}{\multirow{2}{*}{\textbf{Raw feature}}}&\multicolumn{10}{c|}{Subtasks}&\multirow{2}{*}{Avg.}\\
		&&1     & 2     & 3     & 4     & 5     & 6     & 7     & 8     & 9     & 10&\\
		\midrule[0.5pt]
		\multirow{7}{*}{\textit{Single-layer}}&1& 78    & 96    & 94    & 100   & 96    & 98    & 62    & 90    & 68    & 88    & 87.6 \\
		&5&    82    & 94    & 84    & 98    & 94    & 96    & 68    & 94    & 66    & 90    & 86.6 \\
		&9&    94    & 94    & 84    & 94    & 90    & 98    & 90    & 90    & 74    & 90    & \underline{\textit{89.8}\textsuperscript{*}} \\
		&13&    90    & 94    & 86    & 92    & 86    & 100   & 82    & 96    & 84    & 74    & \underline{\textit{88.4}\textsuperscript{*}} \\
		&17&    82    & 92    & 92    & 96    & 92    & 90    & 66    & 72    & 62    & 86    & 84.4 \\
		&21&    78    & 94    & 98    & 90    & 68    & 92    & 66    & 94    & 78    & 88    & 83.2 \\
		&24&    84    & 96    & 94    & 94    & 94    & 100   & 64    & 88    & 56    & 88    & 85.8 \\
		\midrule[0.5pt]
		\textit{All-layer}& 1--24&    92    & 98    & 96    & 100   & 84    & 94    & 76    & 96    & 84    & 86    & \textbf{90.6} \\
		\bottomrule[1pt]
	\end{tabular}
\end{table}%

\begin{table}[!htbp]
	\centering
    \small
	\caption{The specific performance of different layers of ActionQuery features on subtasks. For task instructions, please see Figure \ref{Figure_LIBERO} in Appendix \ref{Appendix_LIBERO_benchmark}. \textbf{Bold} represents the best performance of the average success rate. \underline{\textit{Italics}\textsuperscript{*}} represents the suboptimal performance of the average success rate.}\label{TableC2}%
	\begin{tabular}{cc|cccccccccc|c}
		\toprule[1pt]
		\multicolumn{2}{c|}{\multirow{2}{*}{\textbf{ActionQuery feature}}}&\multicolumn{10}{c|}{Subtasks}&\multirow{2}{*}{Avg.}\\
		&&1     & 2     & 3     & 4     & 5     & 6     & 7     & 8     & 9     & 10&\\
		\midrule[0.5pt]
		\multirow{6}{*}{\textit{Single-layer}}&1&    28    & 50    & 98    & 96    & 80    & 92    & 76    & 94    & 78    & 90    & 78.2 \\
		&13&    16    & 52    & 98    & 94    & 94    & 100   & 66    & 98    & 62    & 86    & 76.6 \\
		&17&    90    & 94    & 86    & 88    & 82    & 100   & 74    & 100   & 58    & 96    & 86.8 \\
		&21&    88    & 92    & 94    & 98    & 92    & 92    & 70    & 96    & 72    & 94    & 88.8 \\
		&23&    92    & 98    & 94    & 96    & 96    & 100   & 70    & 98    & 72    & 82    & \underline{\textit{89.6}\textsuperscript{*}} \\
		&24&    92    & 88    & 100   & 98    & 90    & 96    & 74    & 98    & 84    & 82    & \underline{\textit{90.2}\textsuperscript{*}} \\
		\midrule[0.5pt]
		\textit{All-layer} &1--24&    92    & 94    & 96    & 98    & 100   & 98    & 76    & 98    & 78    & 96    & \textbf{92.6} \\
		\bottomrule[1pt]
	\end{tabular}%
\end{table}%

In Table \ref{TableC1}, the middle-layer Raw features generally outperform other-layer Raw features. In Table \ref{TableC2}, deep-layer ActionQuery features generally perform better than shallow-layer ActionQuery's. In addition, in Table \ref{TableC1} and Table \ref{TableC2}, the performance of all layers is the best. Therefore, in VLA-Adapter, we use the Raw feature and ActionQuery feature of all layers as conditions for Policy.

\vspace{3ex}

\section{Performance on LIBERO Subtasks}\label{AppendixD}
\renewcommand{\thefigure}{D\arabic{figure}}
\renewcommand{\theHfigure}{D\arabic{figure}}
\setcounter{figure}{0}

\renewcommand{\thetable}{D\arabic{table}}
\renewcommand{\theHtable}{D\arabic{table}}
\setcounter{table}{0}

In this section, we demonstrate the performance of VLA-Adapter on 40 (4$\times$10) subtasks on the LIBERO benchmark \citep{LIBERO-2023}. The detailed performance is shown in Table \ref{TableD1}.

\begin{table}[!htbp]
	\centering
	\small
	\caption{The specific performance of VLA-Adapter on the 40 subtasks of four LIBERO \citep{LIBERO-2023} suites. For detailed task instructions, please see Figure \ref{Figure_LIBERO} in Appendix \ref{Appendix_LIBERO_benchmark}.}\label{TableD1}%
	\begin{tabular}{c|cccccccccc|c}
		\toprule[1pt]
		LIBERO & 1     & 2     & 3     & 4     & 5     & 6     & 7     & 8     & 9     & 10    & \textbf{Avg. $\uparrow$} \\
		\midrule[0.5pt]
		Spatial &  98.0     &   100.0    &   100.0    & 90.0      &   96.0    &  100.0     &   100.0    &    100.0   &    98.0   &    96.0   & 97.8 \\
        
		Object & 98.0  & 98.0  & 100.0  & 100.0  & 98.0  & 100.0 & 98.0  & 100.0 & 100.0  & 100.0 & 99.2 \\
        
		Goal  &   92.0    &    100.0   &   98.0    &    96.0   &     100.0  &    98.0   &   94.0    &   100.0    &  98.0     &   96.0    & 97.2 \\
        
		Long  & 96.0&96.0& 100.0 &98.0&100.0&100.0& 84.0& 96.0& 84.0& 96.0&95.0 \\
		\bottomrule[1pt]
	\end{tabular}%
\end{table}%

\vspace{5ex}

\section{Setup Details of CALVIN Simulation Benchmark}\label{AppendixCALVIN}
\renewcommand{\thefigure}{E\arabic{figure}}
\renewcommand{\theHfigure}{E\arabic{figure}}
\setcounter{figure}{0}

\renewcommand{\thetable}{E\arabic{table}}
\renewcommand{\theHtable}{E\arabic{table}}
\setcounter{table}{0}

\begin{figure}[!htbp]
	\centering
	\includegraphics[width=0.99\textwidth]{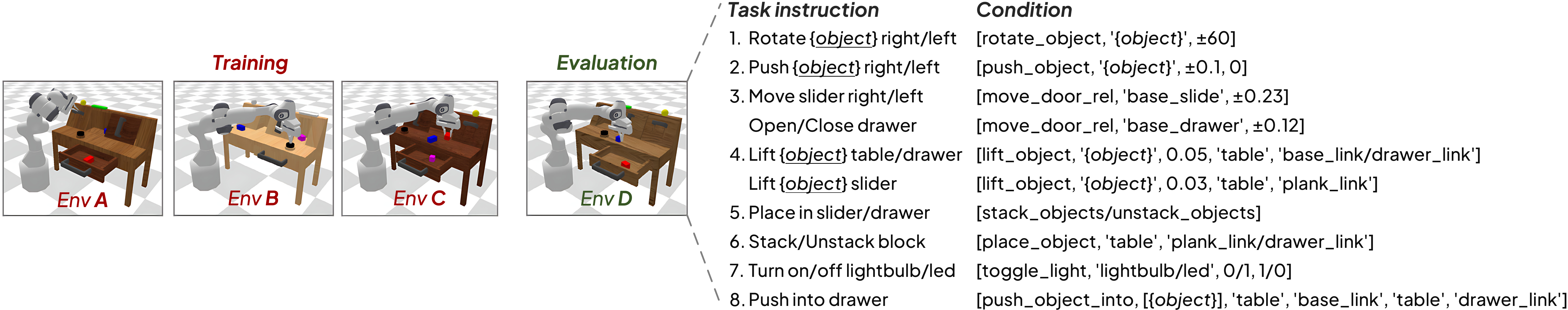}	
	\caption{The example and task completion conditions on the CALVIN ABC$\to$D.}\label{Figure_CALVIN_Bench}
\end{figure}

\paragraph{Benchmark.} We used the CALVIN ABC$\to$D \citep{CALVIN-2022} to evaluate the performance on the zero-shot generalization tasks. CALVIN consists of four environments (Env A, B, C, and D). ``ABC$\to$D" means it trains on Env A, B, and C and evaluates on Env D. These environments collectively include over two million human demonstration trajectories totaling approximately six hours. The CALVIN benchmark contains 34 different subtasks. By screening the combination of five consecutive subtasks, 1,000 unique instruction chains with rationality and diversity are finally generated. In each instruction chain, the agent needs to complete five subtasks in sequence, and can only proceed to the next subtask after successfully completing the current subtask. The benchmark aims to evaluate generalization capabilities and task execution performance under diverse conditions. Examples of each environment and the task instructions are shown in Figure \ref{Figure_CALVIN_Bench}.

In the CALVIN ABC$\to$D benchmark, we use third-person images (resolution 224$\times$224$\times$3, RGB) and Gripper images (resolution 84$\times$84$\times$3, RGB) as visual input. The task instruction is first constructed as a prompt in a specific format: ``\texttt{In: What action should the robot take to \{Task instruction\}?\textbackslash nOut:}", and then input into the VLM module together with the image information. Its output is a 7-dimensional action vector, which is used to control the 7-DOF Franka Emika Panda simulated robot arm to perform the corresponding action sequence.

\vspace{5ex}
\section{Supplementary Details of Training and Hyperparameters}\label{AppendixE}
\renewcommand{\thefigure}{F\arabic{figure}}
\renewcommand{\theHfigure}{F\arabic{figure}}
\setcounter{figure}{0}

\renewcommand{\thetable}{F\arabic{table}}
\renewcommand{\theHtable}{F\arabic{table}}
\setcounter{table}{0}

\subsection{Training Details}\label{AppendixE1}
During VLA-Adapter training, we use the AdamW \citep{AdamW-2019} optimizer and LoRA scheme \citep{LoRA-2022}. To ensure the stability of training, the learning rate is set to 1e-4, and the cosine-annealing scheduler with warm-up steps is used. Our batch size is set to 16.

\begin{table}[!htbp]
	\centering
	\caption{The detail settings of Training.}\label{TableE1}%
	\begin{tabular}{c|c}
		\toprule[1pt]
		Setting & value \\
		\midrule[0.5pt]  
            Batch size & 16 \\
		Max training step & 150,000 \\
		Learning rate & 1e-4 \\
		Warmup step & 10\% \\
		\bottomrule[1pt]
	\end{tabular}%
\end{table}%

\FloatBarrier
\subsection{Hyperparameter Details}\label{AppendixE2}
We list the hyperparameters of VLA-Adapter. Their corresponding values are shown in Table \ref{TableE2}.

\begin{table}[!htbp]
	\centering
	\caption{Specific hyperparameters of VLA-Adapter and their corresponding values.}\label{TableE2}%
	\begin{tabular}{c|c}
		\toprule[1pt]
		Backbone & Qwen2.5-0.5B\\
		Layer ($\tau$ / $M$) & 24  \\
		Number of ActionQuery & 64  \\
		Hidden size & 896   \\
		Attention head & 8      \\
		Action chunk ($H$) & 8\\
		Intermediate layers of VLM & 1--24\\
		\midrule[0.5pt]
		Total trainable parameters of Policy  &97.3M\\
            Total trainable parameters of VLA-Adapter    &197.2M\\
		\bottomrule[1pt]
	\end{tabular}%
\end{table}%

\FloatBarrier
\section{Execution Examples}\label{AppendixF}
\renewcommand{\thefigure}{G\arabic{figure}}
\renewcommand{\theHfigure}{G\arabic{figure}}
\setcounter{figure}{0}

\renewcommand{\thetable}{G\arabic{table}}
\renewcommand{\theHtable}{G\arabic{table}}
\setcounter{table}{0}

\begin{figure}[!htbp]
	\centering
	\includegraphics[width=0.99\textwidth]{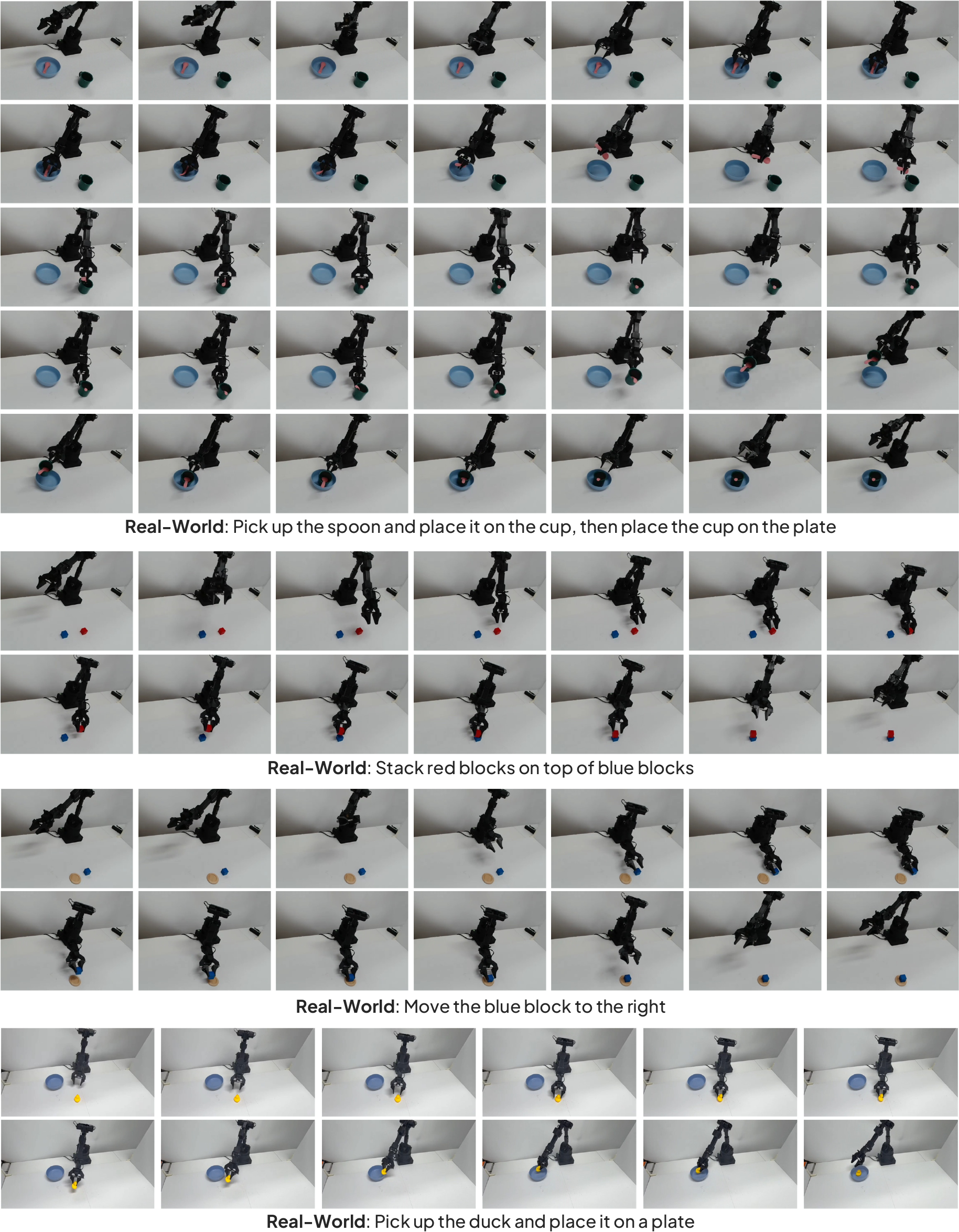}	
	\caption{Execution example on the real-world tasks.}\label{FigF1}
\end{figure}

\begin{figure}[!htbp]
	\centering
	\includegraphics[width=0.99\textwidth]{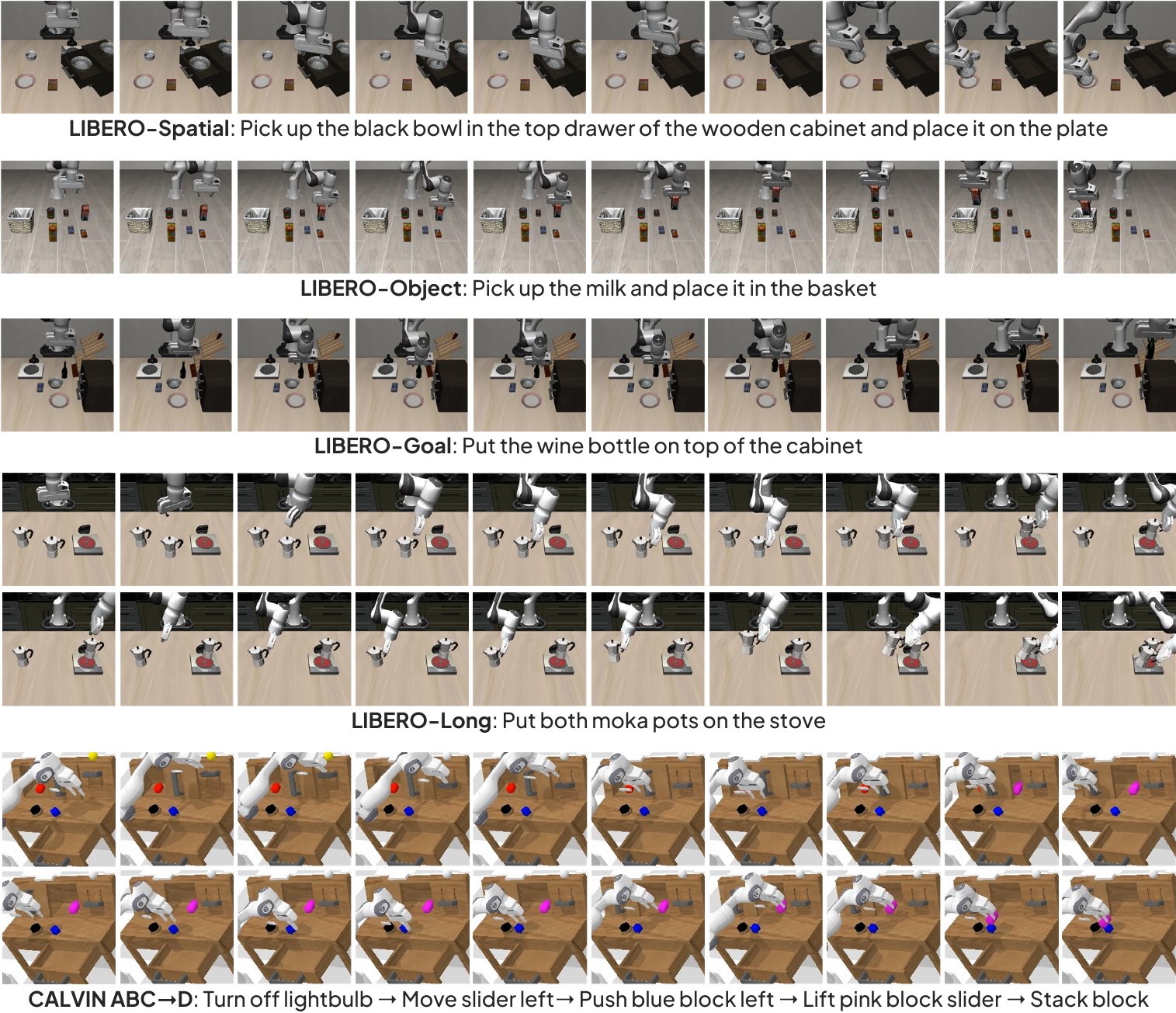}	
	\caption{Execution example on the LIBERO and CALVIN ABC$\to$D tasks.}\label{FigF2}
\end{figure}

We provide some execution examples, please see Figure \ref{FigF1} and Figure \ref{FigF2} for details. 

\subsection{Real-World Examples}\label{AppendixF1}

These include long-horizon tasks: ``Pick up the spoon and place it on the cup, then place the cup on the plate" and short-horizon tasks: ``Stack red blocks on top of blue blocks", ``Move the blue block to the right", and ``Pick up the duck and place it on a plate". The settings of real-world experiments are shown in Section \ref{sec44}.

\FloatBarrier

\newpage
\subsection{Simulation Examples}\label{AppendixF2}

\FloatBarrier

\section{Effectiveness Analysis of Frozen Backbone}\label{AppendixG}
\renewcommand{\thefigure}{H\arabic{figure}}
\renewcommand{\theHfigure}{H\arabic{figure}}
\setcounter{figure}{0}

\renewcommand{\thetable}{H\arabic{table}}
\renewcommand{\theHtable}{H\arabic{table}}
\setcounter{table}{0}

Section \ref{sec41} compares the effectiveness of the frozen backbone. The results show that OpenVLA-OFT does not work. Although it also uses learnable tokens, it is implemented (line 620 in \footnote{\url{https://github.com/moojink/openvla-oft/blob/main/prismatic/extern/hf/modeling_prismatic.py}}):

{\small
	\begin{verbatim}
	# === Handle Multimodal Forward ===
	elif (input_ids.shape[0] == pixel_values.shape[0]) or (inputs_embeds.shape[0] 
                          == pixel_values.shape[0]):
    ...
    # Process action embeddings
    if noisy_actions is not None:
        ...
    else:
        # Replace the embeddings of the action tokens with zeros
        # (Later on, the positional embeddings will be added to them)
        all_actions_mask = all_actions_mask.unsqueeze(-1) 
        input_embeddings = input_embeddings * ~all_actions_mask
	\end{verbatim}
}

The tokens added in the ``else:" (L1 architecture) are input to the VLM in the form of a mask. It is initially all zeros, and when the VLM backbone is frozen, it is not trained. Our ActionQuery is:

{\small
	\begin{verbatim}
	# === Handle Multimodal Forward ===
	elif (input_ids.shape[0] == pixel_values.shape[0]) or (inputs_embeds.shape[0] 
                          == pixel_values.shape[0]):
    ...
    # Process action embeddings
    if noisy_actions is not None:
        ...
    else:
        action_queries = self.action_queries.weight  # (1, h)
        action_queries = action_queries.view(1, action_queries.shape[0], 
            action_queries.shape[1]).repeat(input_embeddings.shape[0], 1, 1)
                  
        all_actions_mask = self._process_action_masks(labels)
        input_embeddings = self._replace_input_embeddings(input_embeddings, 
                                                          all_actions_mask, 
                                                          action_queries)
	\end{verbatim}
}

\begin{figure}[!htbp]
	\centering
	\includegraphics[width=0.99\textwidth]{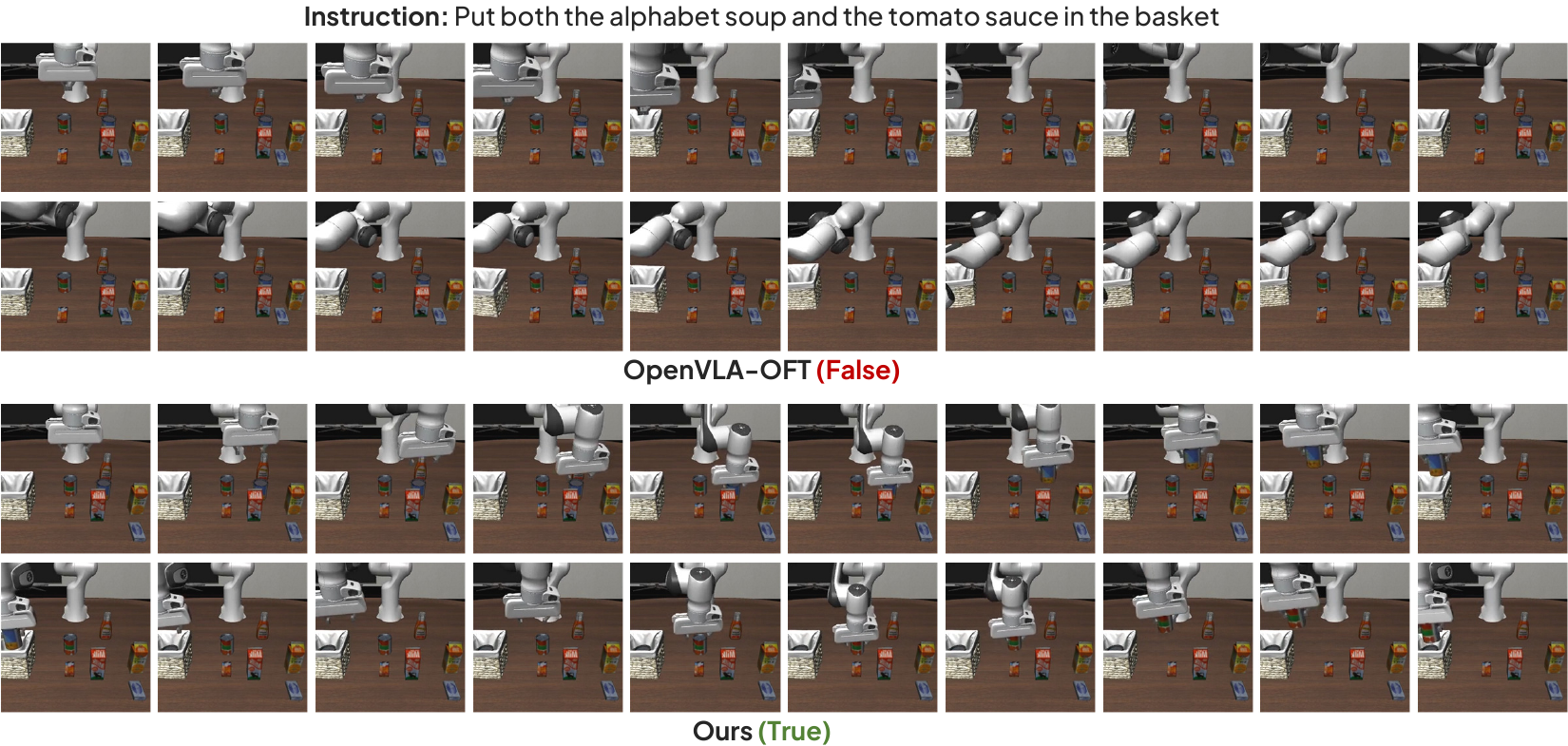}	
	\caption{Execution example when the backbone is frozen.}\label{FigG1}
\end{figure}

Instead of inputting the VLM in the form of a mask, essentially learnable tokens (or multiple tokens) that is inserted into the specified position in the sequence and participate in attention. Therefore, when the VLM is frozen, the VL information is indeed not trained. Still, ActionQuery is not an original part of the VLM, and its parameters can be learned from scratch, so the VLA-Adapter still works. Below, we give examples of OpenVLA-OFT and VLA-Adapter, as shown in Figure \ref{FigG1}.

\vspace{11ex}

\section{Design Details and Performance of VLA-Adapter-Pro}\label{AppendixI}
\renewcommand{\thefigure}{I\arabic{figure}}
\renewcommand{\theHfigure}{I\arabic{figure}}
\setcounter{figure}{0}

\renewcommand{\thetable}{I\arabic{table}}
\renewcommand{\theHtable}{I\arabic{table}}
\setcounter{table}{0}

To achieve extreme lightweightness in VLA-Adapter, we shared the projection layer, using the same projection layer for all three attention matrices. This resulted in a low parameter count of 97MB. 

In the VLA-Adapter-Pro version, we separated the projection layers for the three attention matrices, allowing different channels to learn differentiated representations. In this case, the parameter count was 207MB. Furthermore, VLA-Adapter-Pro adds Rotary Position Embedding to QK, making the cross-attention more sensitive to position information and more suitable for action generation.

Next, we give the key architecture codes of VLA-Adapter-Pro, as shown below:

{\small
\begin{verbatim}
class MLPResNetBlock_Pro(nn.Module):
    """One MLP ResNet block with 
    separate projections for self, adapter, task 
    + 
    RoPE, now without FiLM modulation."""

    def __init__(self, dim, num_heads=8):
        super().__init__()
        self.dim = dim
        self.num_heads = num_heads
        self.head_dim = dim // num_heads

        self.ffn = nn.Sequential(
            nn.LayerNorm(dim),
            nn.Linear(dim, dim),
            nn.ReLU(),
            )

        # Q (from x only)
        self.q_proj = nn.Linear(dim, dim)

        # Self-Attention: K, V
        self.k_self = nn.Linear(dim, dim)
        self.v_self = nn.Linear(dim, dim)

        # Adapter cross-attention: K, V
        self.k_adapter = nn.Linear(dim, dim)
        self.v_adapter = nn.Linear(dim, dim)

        # Task cross-attention: K, V
        self.k_task = nn.Linear(dim, dim)
        self.v_task = nn.Linear(dim, dim)

        self.o_proj = nn.Linear(dim, dim)

        # gating
        self.gating_factor = nn.Parameter(torch.zeros(1))

        # RoPE
        self.rope = RotaryPositionEmbedding(self.head_dim)

        # ---- FiLM ----
        # # FiLM is useless; to avoid conflict with chkpt, it can be kept as is for now.
        self.film_gen = nn.Sequential(
            nn.Linear(dim, dim * 2), 
            )

    def apply_film(self, x, gamma, beta):
        """FiLM: per-channel modulation"""
        return gamma.unsqueeze(1) * x + beta.unsqueeze(1)

    def forward(self, x, h_a=None, h_t=None, p=None):
        """
        h_a: adapter tokens
        h_t: task tokens
        p:   possible conditioning vector (for FiLM)
        """
        g = self.gating_factor
        ratio_g = torch.tanh(g)

        # concat h_a and p
        h_adapter = torch.cat((h_a, p),dim=1)

        h_task = h_t
        B, T, C = x.shape
        K_a = h_adapter.size(1) if h_a is not None else 0
        K_t = h_task.size(1) if h_task is not None else 0

        # Q
        q_1 = self.q_proj(x)

        # self tokens
        k_tokens = self.k_self(x)
        v_tokens = self.v_self(x)

        # adapter tokens
        k_adapter = self.k_adapter(h_adapter)
        v_adapter = self.v_adapter(h_adapter)

        # task tokens
        k_task = self.k_task(h_task)
        v_task = self.v_task(h_task)

        # reshape -> multi-head
        def reshape_heads(t, B, L):
            return t.view(B, L, self.num_heads, self.head_dim).transpose(1, 2)

        q_1 = reshape_heads(q_1, B, T)
        k_tokens, v_tokens = reshape_heads(k_tokens, B, T), reshape_heads(v_tokens, B, T)
        k_adapter, v_adapter = reshape_heads(k_adapter, B, K_a), reshape_heads(v_adapter, B, K_a)
        k_task, v_task = reshape_heads(k_task, B, K_t), reshape_heads(v_task, B, K_t)

        # RoPE
        cos_main, sin_main = self.rope(seq_len=T, device=x.device, dtype=x.dtype)
        q_1, k_tokens = apply_rope(q_1, k_tokens, cos_main, sin_main)
        cos_a, sin_a = self.rope(seq_len=K_a, device=x.device, dtype=x.dtype)
        _, k_adapter = apply_rope(k_adapter, k_adapter, cos_a, sin_a)     
        cos_t, sin_t = self.rope(seq_len=K_t, device=x.device, dtype=x.dtype)
        _, k_task = apply_rope(k_task, k_task, cos_t, sin_t)

        # attention scores
        attn_scores = [torch.matmul(q_1, k_tokens.transpose(-2, -1))]
        attn_scores.append(torch.matmul(q_1, k_adapter.transpose(-2, -1)))
        attn_scores.append(torch.matmul(q_1, k_task.transpose(-2, -1)) * ratio_g)
        attn_scores = torch.cat(attn_scores, dim=-1) / math.sqrt(self.head_dim)
        attn_weights = torch.softmax(attn_scores, dim=-1)

        # combine V
        v_list = [v_tokens,v_adapter,v_task]
        v_combined = torch.cat(v_list, dim=2)

        output = torch.matmul(attn_weights, v_combined)
        output = output.transpose(1, 2).contiguous().view(B, T, C)
        output = self.o_proj(output)

        # # ---- FiLM ---- 
        # gamma_beta = self.film_gen(p)  # [B, 2C]
        # gamma, beta = gamma_beta.chunk(2, dim=-1)  # [B, C], [B, C]
        # output = self.apply_film(output, gamma, beta)

        # residual + FFN
        x = self.ffn(output + x)
        return x

\end{verbatim}
}

We also present the performance comparison between VLA-Adapter and VLA-Adapter-Pro on 40 subtasks on LIBERO, as shown in Table \ref{TableI1}.

\begin{table}[!htbp]
	\centering
    \setlength{\tabcolsep}{1mm}
	\caption{The specific performance of VLA-Adapter and VLA-Adapter-Pro on the 40 subtasks of four LIBERO \citep{LIBERO-2023} suites.}\label{TableI1}%
	\begin{tabular}{c|cccccccccc|c}
		\toprule[1pt]
		Spatial & 1     & 2     & 3     & 4     & 5     & 6     & 7     & 8     & 9     & 10    & \textbf{Avg. $\uparrow$} \\
		\midrule[0.5pt]
		VLA-Adapter &  98.0     &   \textbf{100.0}    &   100.0    & 90.0      &   96.0    &  100.0     &   100.0    &    100.0   &    98.0   &    96.0   & 97.8 \\
        VLA-Adapter-Pro &  \textbf{100.0}     &   98.0    &   100.0    & \textbf{100.0}      &   \textbf{100.0}    &  100.0     &   100.0    &    100.0   &    \textbf{100.0}   &    \textbf{98.0}   & \textbf{99.6} \\
        \midrule[1pt]
        
		Object & 1     & 2     & 3     & 4     & 5     & 6     & 7     & 8     & 9     & 10    & \textbf{Avg. $\uparrow$} \\
		\midrule[0.5pt]
        VLA-Adapter & 98.0  & 98.0  & 100.0  & 100.0  & 98.0  & 100.0 & 98.0  & 100.0 & 100.0  & 100.0 & 99.2 \\
        VLA-Adapter-Pro & \textbf{100.0}  & \textbf{100.0}  & 100.0  & 100.0  & 98.0  & 100.0 & 98.0  & 100.0 & 100.0  & 100.0 & \textbf{99.6} \\
        \midrule[1pt]
        
		Goal & 1     & 2     & 3     & 4     & 5     & 6     & 7     & 8     & 9     & 10    & \textbf{Avg. $\uparrow$} \\
        \midrule[0.5pt]
        VLA-Adapter&   92.0    &    100.0   &   \textbf{98.0}    &    96.0   &     100.0  &    98.0   &   94.0    &   100.0    &  98.0     &   96.0    & 97.2 \\
        VLA-Adapter-Pro & \textbf{98.0}  &100.0  & 94.0  & 96.0  & 100.0  & 98.0 & \textbf{96.0}  & 100.0 &\textbf{100.0}  & \textbf{100.0} & \textbf{98.2} \\
        \midrule[1pt]
        
		Long  & 1     & 2     & 3     & 4     & 5     & 6     & 7     & 8     & 9     & 10    & \textbf{Avg. $\uparrow$} \\
        \midrule[0.5pt]
        VLA-Adapter&\textbf{96.0}&96.0& \textbf{100.0} &\textbf{98.0}&\textbf{100.0}&100.0& 84.0& 96.0& 84.0& 96.0&95.0 \\
        VLA-Adapter-Pro&92.0&\textbf{100.0}& 98.0 &96.0&94.0&100.0& \textbf{94.0}& \textbf{100.0}& \textbf{90.0}& \textbf{100.0}&\textbf{96.4} \\
		\bottomrule[1pt]
	\end{tabular}%
\end{table}%